\documentclass[acmsmall]{acmart}

\usepackage[boxruled, linesnumbered]{algorithm2e}
\usepackage{xspace}

\AtBeginDocument{%
  \providecommand\BibTeX{{%
    \normalfont B\kern-0.5em{\scshape i\kern-0.25em b}\kern-0.8em\TeX}}}

\setcopyright{acmlicensed}
\copyrightyear{2021}
\acmYear{2021}
\acmDOI{10.1145/1122445.1122456}

\acmJournal{TIIS}
\acmVolume{1} 
\acmNumber{1} 
\acmArticle{1} 
\acmMonth{1}
\acmPrice{15.00}
\acmDOI{10.1145/3484509}

\begin{document}

\title{Tribe or Not? Critical Inspection of Group Differences Using TribalGram}

\author{Yongsu Ahn}
\affiliation{%
  \institution{University of Pittsburgh}
  \city{Pittsburgh}
  \country{United States}}
\email{yongsu.ahn@pitt.edu}

\author{Muheng Yan}
\affiliation{%
  \institution{University of Pittsburgh}
  \city{United States}
  \country{United States}}
  \email{muheng.yan@pitt.edu}

\author{Yu-Ru Lin}
\affiliation{%
  \institution{University of Pittsburgh}
  \city{United States}
  \country{United States}}
  \email{yurulin@pitt.edu}
  
\author{Wen-Ting Chung}
\affiliation{%
  \institution{University of Pittsburgh}
  \city{United States}
  \country{United States}}
  \email{wtchung@pitt.edu}
  
\author{Rebecca Hwa}
\affiliation{%
  \institution{University of Pittsburgh}
  \city{United States}
  \country{United States}}
  \email{hwa@pitt.edu}

\renewcommand{\shortauthors}{Ahn et al.}

\newcommand{\yrl}[1]{{\color{red}{[YRL:#1]}}}
\newcommand{\ysc}[1]{{\color{blue}{[YS:#1]}}}

\newcommand{\ys}{\textcolor{blue}}
\newcommand{\wtc}{\textcolor{orange}}
\newcommand{\yrv}[1]{{\color{purple}{#1}}}

\newcommand{\name}{\texttt{TribalGram}\xspace}
\newcommand{\fname}{AccountG\xspace} 

\definecolor{brandeisblue}{rgb}{0.0, 0.44, 1.0}
\newcommand{\blue}{{\textcolor{brandeisblue}{Blue}}\xspace}
\definecolor{carminepink}{rgb}{0.92, 0.3, 0.26}
\newcommand{\red}{{\textcolor{carminepink}{Red}}\xspace}

\newcommand{\blueclr}{{\textcolor{brandeisblue}{blue}}\xspace}
\newcommand{\redclr}{{\textcolor{carminepink}{red}}\xspace}

\newcommand{\vale}{{\it Valence}\xspace}
\newcommand{\domi}{{\it Dominance}\xspace}
\newcommand{\care}{{\it Care}\xspace}
\newcommand{\fair}{{\it Fairness}\xspace}
\newcommand{\puri}{{\it Purity}\xspace}
\newcommand{\loya}{{\it Loyalty}\xspace}
\newcommand{\auth}{{\it Authority}\xspace}
\newcommand{\tweet}[1]{``\textit{#1}''}

\newcommand{\subgroup}{Subgroup\xspace}

\newcommand{\instanceviewer}{\texttt{Instance Viewer}\xspace}
\newcommand{\scopecontroller}{\texttt{Scope~Controller}\xspace}
\newcommand{\grouptrend}{\texttt{GroupTrend}\xspace}
\newcommand{\evalscope}{\texttt{EvalScope}\xspace}
\newcommand{\variscope}{\texttt{VariScope}\xspace}
\newcommand{\depscope}{\texttt{DepScope}\xspace}
\newcommand{\rationalescope}{\texttt{RationaleScope}\xspace}
\newcommand{\languagescope}{\texttt{LanguageScope}\xspace}
\newcommand{\clusterview}{Cluster View\xspace}

\newcommand{\listview}{List View\xspace}
\newcommand{\generator}{Generator\xspace}
\newcommand{\parallelset}{Parallel Set\xspace}
\newcommand{\sequenceview}{Sequence View\xspace}
\newcommand{\subgroupview}{Subgroup View\xspace}

\renewcommand{\vec}[1]{\mathbf{#1}} 
\newcommand{\mat}[1]{\mathbf{#1}} 
\newcommand{\ran}{\mathbb{R}} 

\newcommand{\rev}[1]{{\color{blue}{#1}}}
\newcommand{\revyongsu}[1]{{\color{blue}{#1}}}
\newcommand{\revys}[1]{{\color{red}{#1}}}

\begin{abstract}
With the rise of AI and data mining techniques, group profiling and group-level analysis have been increasingly used in many domains including policy making and direct marketing. In some cases, the statistics extracted from data may provide insights to a group's shared characteristics; in others, the group-level analysis can lead to problems including stereotyping and systematic oppression. How can analytic tools facilitate a more conscientious process in group analysis? In this work, we identify a set of {\it accountable group analytics} design guidelines to explicate the needs for group differentiation and preventing overgeneralization of a group. Following the design guidelines, we develop {\it\name}, a visual analytic suite that leverages interpretable machine learning algorithms and visualization to offer inference assessment, model explanation, data corroboration, and sense-making. Through the interviews with domain experts, we showcase how our design and tools can bring a richer understanding of ``groups'' mined from the data.
\end{abstract}



\begin{teaserfigure}
\centering
  \includegraphics[width=0.9\columnwidth]{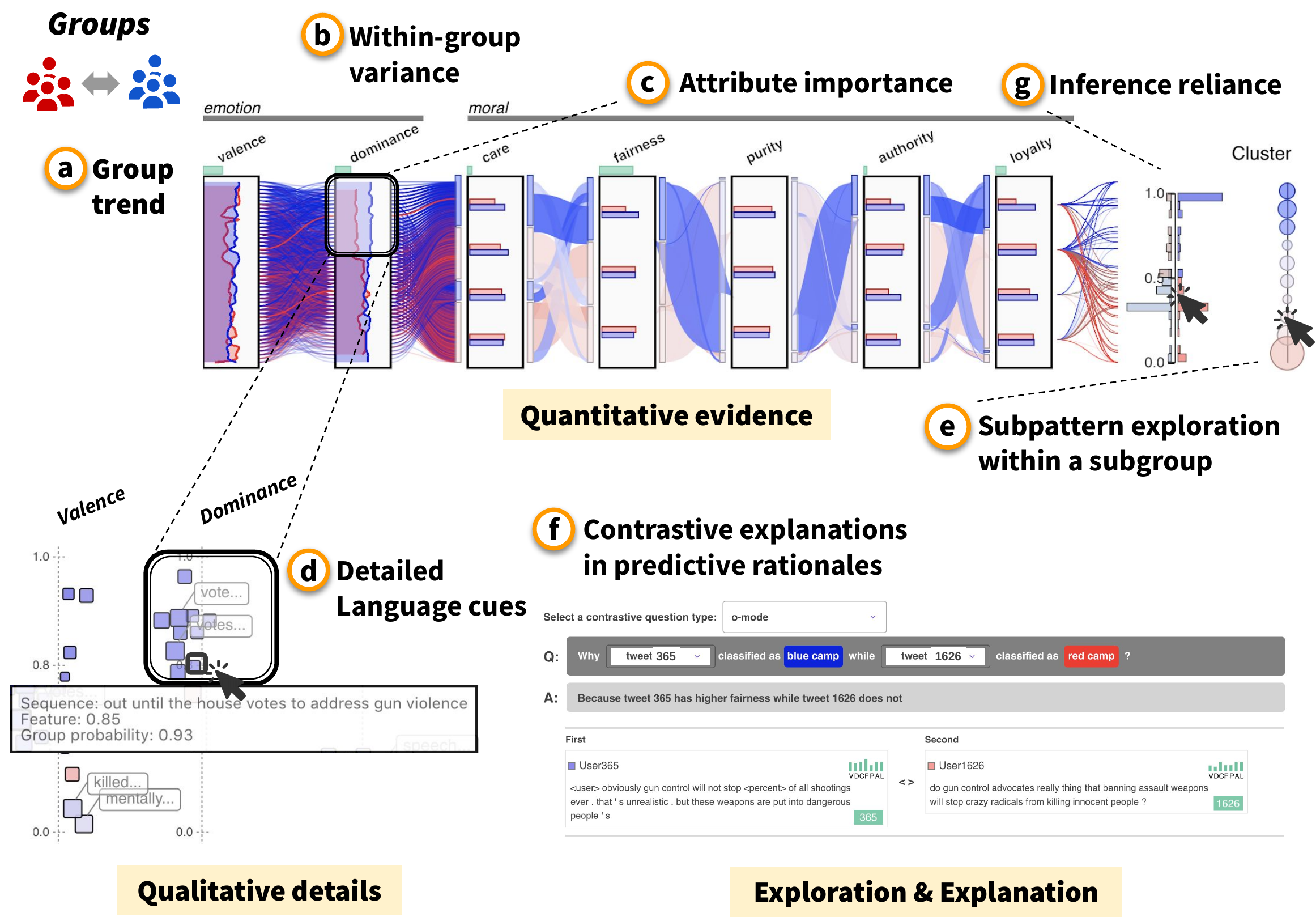}
  \caption{
  \name provides a critical inspection of group differences to facilitate accountable group analytics with a visual analytic suite. It allows users to (a) visually contrast groups' main characteristics (\grouptrend), (b) examine the within-group variance (\variscope), (c) capture group-level attribute importance (\depscope), (d) retrieve qualitative language details (\languagescope), and (e) explore nested patterns and diversity within a group. It also supports users to (g) assess the analytic model quality against ground-truth data whenever available (\evalscope), and (f) obtain the rationale behind the group assignment of any individual instance (\rationalescope).
  }
  \label{fig:teaser}
\end{teaserfigure}

\maketitle

\section{Introduction} \label{sec:introduction}

Group-level analysis plays an important role in social sciences. With the rise of big data, artificial intelligence (AI), and data mining techniques, group analysis has increasingly become a powerful tool in many applications, ranging from policy-making, direct marketing, education, to healthcare.
For example, an important analysis strategy is {\it group profiling}, which extracts and describes the characteristics of groups of people \cite{hildebrandt2006profiling}; it has been commonly used for customized recommendations to overcome sparse and missing personal data \cite{dehghani2016generalized}. 
The same strategy is also used for mining social media, educational, and healthcare data to understand the shared characteristics of online communities or student/patient cohorts \cite{zhang2015iterative,krause_supporting_2016,cao2015g}. While it may help to support public and private services or product creations that are better tailored to different communities, group profiles resulted from mathematical inference are typically not valid for every individual regarded as a member in the group (this is known as {\it non-distributive} group profiles) \cite{hildebrandt2006profiling}. The shared group characteristics extracted from data can have social ramifications such as stereotyping, stigmatization, or lead to pernicious consequences in decision making because individuals might be judged by group characteristics they do not posses \cite{custers2004power,mendoza2017right,gdpr}.

Given the extensive use of group-level analysis, researchers have begun to examine its ramifications. Prior work has dealt with issues ranging from model evaluation to contextualized group interpretability \cite{zhang2015iterative,krause_supporting_2016,cao2015g}, and recent progress in interpretable Machine Learning (iML) and eXplainable Artificial Intelligence (XAI) have brought much attention and methods to diagnose and produce more explainable models \cite{adadi2018peeking,guidotti2018survey}. However, explainable models alone will not solve this problem. Even if the algorithms or models are completely accurate, understandable and trustworthy, the risk of overgeneralization of group characteristics still exists, especially when most group-level analytic seem to highlight such generalization backed up by mathematical inference -- rather than to examine groups more conscientiously. 

While there is a clear need to provide analytics to enable users to characterize groups of interest, at the same time, the appropriate assessment must also be provided to mind the potential overgeneralization of group characteristics. How can analytic tools facilitate a more conscientious practice in such analyses? We address this larger question by answering two sub-questions: (1) What are the design requirements for developing such analytic tools? (2) What technical methods and concepts are needed to create group-level analytic tools that meet these requirements? We first conduct multiple rounds of requirement interviews with domain experts to identify a set of design guidelines for {\it accountable group analytic}. Following the design guidelines, we develop {\it\name}, a visual analytic suite that leverages interpretable machine learning algorithms and visualization to enable inference assessment, model explanation, data corroboration and sense-making analysis. 

As an example, suppose a data journalist wants to understand social media users' conversations about  ``gun control" issues. One possible question might be: ``{\it how do conservatives and liberals talk differently on the gun issues?}'' Using \name and carefully curated social media samples, the journalist may begin with a summary view of the major sociolinguistic differences between the two camps -- for example, after a mass shooting event, one camp might {\it generally} post with a more positive tone, expressing more solidarity and care, or blaming suspects or legislation, compared to the other (Fig.~\ref{fig:teaser}a:~\grouptrend). But {\it how do such sociolinguistic differences manifest in users' communications?}  \name allows the journalist to look for quotes from the users' posts that support the characterization of sociolinguistic attributes (Fig.~\ref{fig:teaser}d:~\languagescope), or see how a camp itself may exhibit diverse patterns in terms of any attribute (Fig.~\ref{fig:teaser}b,c:~\variscope, \depscope). The journalist can see the extent to which those attributes explain the political leaning of users (Fig.~\ref{fig:teaser}c:~\depscope). Moreover, using model explanation tool (Fig.~\ref{fig:teaser}f:~\rationalescope), the journalist can see {\it why} a model might categorize an individual as more conservatives than liberals (or vice versa) given his/her sociolinguistic tendency.

Our key contributions include: (1) We identify a set of design guidelines for creating group-level analysis tools to facilitate a critical inspection of groups. This is the first work that explores the design issues concerning non-distributive group profiles and the risk of group overgeneralization. (2) We propose a new visual analytic toolkit, \name, that incorporates (a) a suite of visual analytic components to help quantitatively and qualitatively inspect the shared and varied characteristics of groups, and (b) interpretable machine learning algorithms, including multi-task predictive models and contrastive explanatory models, to identify qualitative evidence and rationale for individuals regarded as group members. Despite that our \name is developed for a specific scenario -- the inspection of ideological groups using social media data, the visual and algorithmic components can be applied to the analysis of social groups in many domains.

To evaluate the utility of our proposed design, we conduct interview studies with experts from three different domains. Our study suggests that the critical inspection design in \name not only enables domain experts to identify distinctions between groups but also allows them to search and test new hypotheses, explore borderline and edge cases, and find qualitative cues for the group characteristics, which together leads to a richer understanding of the data being grouped.

\section{Related Work} \label{sec:related-work}

Group-level analytics with data about people involves data mining techniques to identify groups using observed individual features, with a variety of visual analytic techniques proposed to help users understand the identified groups. This has been joined by the recent progress in explainable AI or interpretable machine learning, with a goal to enable human users to understand the data-driven decision-making processes. We briefly review works that are the most related to ours in the three areas.

\subsection{Group identification from observed data}
Online user-generated contents such as users' social media posts and activities have been increasingly used in many domains for understanding the characteristics of different sociodemographic groups. Groups can be characterized in a deductive or inductive manner. The deductive approach starts with certain hypotheses (e.g., groups are different in terms of some attributes) and uses top-down or supervised learning methods to find groups with certain attributes \cite{phillips2017using}. On the other hand, unsupervised learning methods, such as community detection, are used to group individuals with similar patterns without presumptions of the group characteristics \cite{lin2009metafac,tang2011group}. In the context of social media data where the hypothesized attributes are not readily available, text mining techniques have been utilized to extract more explicit attributes into theorized constructs to analyze individuals, with the use of sociolinguistic features, words and hashtags \cite{an_greysanatomy_nodate, carpenter_real_2017, conover2011political} and other social media behavior \cite{bamman_gender_2014, filippova_user_nodate, wood-doughty_how_2017}. Other types of data, such as demographic metadata \cite{sloan_who_2015} or network structure \cite{rao_classifying_2010}, may be used in addition to the user-generated text based features, such as linguistic occurrences based on LIWC dictionary \cite{fink_inferring_nodate}, words \cite{burger_discriminating_nodate, jensen2012political}, or embeddings \cite{hovy_demographic_2015, demszkyAnalyzingPolarizationSocial2019}, to algorithmically identify target groups, including predicting gender \cite{fink_inferring_nodate, sap_developing_2014}, age \cite{guimaraes_age_2017, lopez_predicting_2017} and political preferences \cite{cohen_classifying_nodate, volkova_inferring_2014, volkova2015online}.

Many of the features extracted from user-generated text, such as the use of specific words and phrases, may not have a theoretical justification, which hinders the interpretation of the generated groups -- for example, whether a group has characteristics as previously hypothesized. Several works have especially focused on extracting the interpretable, theoretically-grounded, or higher-level differences between groups, such as sentiment and emotion \cite{volkova_predicting_2015, volkova_inferring_2016}, personality \cite{golbeck_predicting_nodate, wei_beyond_2017}, or moral values \cite{kalimeri_predicting_2019}. For example, Svitlana et al. \cite{volkova_inferring_2016} examined how users' emotional dimensions are different from their social contacts and found such differences are discriminative in predicting socio-demographic groups such as ethnicity and gender. Kalimeri et al. \cite{kalimeri_predicting_2019} analyzed the psychometric questionnaires and web browsing behavior to infer moral trait and human values.

Such inferred, higher-level variables may create new challenge in interpretation as the variables may not correspond to a particular example in the raw text. In this work, we introduce an analytic tool that bridges the high-level, theoretically-grounded attributes and the low-level, language instances that enables users to find evidence directly from user-generated text for how the inferred attributes reflect the concept it purports to measure.

\subsection{Visual analytics for group difference}
Visual analytics has been designed to support group-level analysis in many application domains. Typical analysis tasks include 
(1) selecting a group of people that meet certain criteria (e.g., constructing patient cohorts to support cohort analysis or intervention design) \cite{klemm_interactive_2014, kwon_retainvis_2019, zhang2015iterative,krause_supporting_2016}, (2) identifying certain socio-demographic groups, or (3) clustering people using observed data over multivariate attributes \cite{cavallo_clustrophile_2019, kwon_clustervision_2018, pearlman2007visualizing, pham2010visualization, pham2014visualization}. For example, Klemm et al. \cite{klemm_interactive_2014} visualized the averaging medical images of groups with respect to socio-demographic metadata or clustering analysis. Interactive techniques are introduced in sub-setting cohorts who have certain temporal constraints \cite{krause_supporting_2016, zhang2015iterative}. Krause et al. \cite{krause_supporting_2016} proposed a visual filtering layout for specifying cohorts, while CAVA \cite{zhang2015iterative} proposed a flow visualization where a subset of cohorts can be aggregated and divided by features over discrete timepoints.

Although most visualization works support the interactive exploration of group patterns, fewer tools focus on supporting both the group classification and group differences. One such work is DemographicVis \cite{dou_demographicvis_2015}, which proposed a visual text analytics system for exploring the group inference task and the underlying topical patterns of user-generated contents across demographic groups. While this work showed that a group classification model may be improved by adding linguistic and topical features, the process of group inference and the identified important features are not interpretable to the end-users. Some of studies on visual analytic tools with the classification task investigate the interative subgrouping but with the focus of discovering bias and model failure \cite{ahn2019fairsight, cabrera2019fairvis, gleicher2020boxer, wexler2019if, chung2019slice}. In this work, we introduce a new visual analytic suite that enables not only the exploration of group patterns but also the reasoning of important attributes and of individuals in a group.

\subsection{Deep learning and interpretable machine learning}
Deep neural networks has been shown to help with identifying group differences, such as sentiment \cite{can2018multilingual,chen2018twitter}, emotion \cite{abdullah2018sedat,gupta2017sentiment}, affect \cite{abid2019sentiment}, and morality \cite{rezapour2019enhancing}. The flexible architectures in deep neural networks make it effective in taking diverse types of data input ranged from structured variables, text \cite{sundermeyer2012lstm}, to image \cite{kalchbrenner2014convolutional}, into predictive features. For example, Recurrent Neural Networks (RNN) \cite{sundermeyer2012lstm}  can take a sequence of tokens (words or phrases) from natural languages as input, and recursively encode and transformed the textual input to a latent space while keeping the language structures. However, the transformed features resulted from the learning layers are often difficult to interpret. Thus, how to enhance the interpretability of neural network models has drawn considerable attention. One example technique is the use of the attention mechanism, first introduced by Cho et al. \cite{cho2014learning} on RNNs for machine translation. In this approach, the recursively encoded tokens are weighted in the learning of the model, so that contributions of individual token inputs to the model prediction are discoverable through the weights. The attention mechanism has been extended in different contexts. Xu et al. \cite{xu2015show} used attention to identify informative pixel areas from images. Hermann et al. \cite{hermann2015teaching} used attention for the extraction of important sections from text paragraphs and Bahdanau et al. \cite{bahdanau2014neural} used attention for word alignment in machine translation.

In this work, we introduce a new multi-task prediction task that leverages RNN architecture and attention mechanism with a joint objective to simultaneously predict group labels and attributes. As a result, the attention mechanism enables the finding of language cues as evidence to support the interpretation of groups' sociolinguistic attributes.

\section{Design Guidelines \& Tasks} \label{sec:goalsandtasks}

\begin{table}[]
\resizebox{\textwidth}{!}{%
\begin{tabular}{@{}lllll@{}}
\toprule
 &
  \textbf{Expertise} &
  \textbf{Goal of analysis} &
  \textbf{Workflow} &
  \textbf{Challenges and Limitations} \\ \midrule
\textbf{Expert A} &
  \begin{tabular}[c]{@{}l@{}}Online movements \\ and campaigns\end{tabular} &
  \textit{\begin{tabular}[c]{@{}l@{}}"How do groups hold \\ different emotional and \\ moral attitudes towards \\ social issues and capture \\ a concrete evidence \\ from their utterances?"\end{tabular}} &
  \begin{tabular}[c]{@{}l@{}}Count the word occurrences in R;\\ Fit multiple models with \\ filtered words for each attribute;\end{tabular} &
  \begin{tabular}[c]{@{}l@{}}Finding anecdotal evidence \\ for complex psychological \\ dimensions is not feasible, \\ or it is shallow when available\end{tabular} \\
\textbf{Expert B} &
  \begin{tabular}[c]{@{}l@{}}Team \\ communication\end{tabular} &
  \textit{\begin{tabular}[c]{@{}l@{}}"How do demographic groups \\ form the faultline with respect \\ to team members' attributes \\ within or across the groups?"\end{tabular}} &
  \begin{tabular}[c]{@{}l@{}}Run a linear regression with STATA; \\ Do clustering analysis with Tableau;\\ Examine subgroup members \\ in spreadsheet;\end{tabular} &
  \begin{tabular}[c]{@{}l@{}}Interaction between deductive \\ and inductive analysis is \\ inconvenient; (i.e., how is a \\ subgroup characterized \\ (inductive) or attributed \\ (deductive) to each attribute?\end{tabular} \\ \bottomrule
\end{tabular}%
}
\caption{Examples of the pilot study result.}
\label{tbl:pilot_study}
\end{table}

In this section, we propose a set of design guidelines of group-level analytic tools for facilitating a conscientious practice of analyzing group characteristics. To formulate the design of a group analysis tool, we collect the feedback and thoughts from potential users in using existing tools. We assume that such tools will be used by data analysts and experts who conduct a group-level analysis in different domains. In order to understand the current practice and goal of group analysis and the challenges from it, we conducted the pilot study with domain experts and identified the concerns and limitations of the group-level analysis in their workflow. Based on the findings of the pilot study, our design guideline, which consists of three goals and six specific tasks, was formulated by compiling common aspects across the interviewees’ comments. In the following paragraphs, we describe the process of deriving design guidelines in detail.

\textbf{Pilot study.} Five interviewees were chosen from a variety of domains and specializations, ranging from data science, psychology, education, language, to anthropology. All these domain experts had prior experiences with human group research using digital trace data. Each interview lasted about an hour, in the form of a semi-structured interview. During the interview, we engaged the domain experts to consider a simple scenario of using structured and unstructured data to conduct group analyses, encouraged them to apply their current practices, and reflect on the limitations and concerns of the analytic tools they currently used. We organized the interview session with three primary questions to facilitate the interviewees’ thinking process:

\begin{itemize}
    \item \textbf{Expertise/Workflow}: “What is your expertise and in what way typically do you get insights on such analysis?”
    \item \textbf{Goal of analysis}: “What insights do you anticipate to find out?”
    \item \textbf{Challenges/Limitations}: “Despite your current practice of group analysis with available tools, how does it fail to meet your needs?”
\end{itemize}

While interviewees reported a set of goals and challenges/limitations in their own context of group analysis as shown in Table \ref{tbl:pilot_study}, we were able to find that they had experienced similar difficulties in their goals and tasks. For example, a typical analysis goal of three interviewees is not only to test their hypothesis over the group/subgroup characteristics but also to find qualitative evidence to back up their test statistics. As summarized in Table \ref{tbl:pilot_study}, many existing tools they currently used in their workflow, such as STATA, R, Tableau, or simple software for dealing with spreadsheet, did not fully support the range of group analytic function -- from overviewing group characteristics, running a regression/prediction model, to identifying subgroup patterns across attributes.

After all interviews were conducted, we compiled a list of pairs of goal and challenges for each interviewee (within-interview), and grouped common facets of pairs (e.g., (goal) \textit{capture both quantitative and qualitative evidence} - (challenge) \textit{hard to capture it with respect to each attribute}) shared by interviewees (between-interviews). Based on the result of the interview study, we identified six concerns that were commonly mentioned by interviewees (denoted as \textbf{C}):

\begin{itemize}
    \setlength{\itemsep}{0.2pt}
    \item[\textbf{C1.}] Hard to capture the overview of group differences.
    \item [\textbf{C2.}] No links or traces between deductive and inductive analysis.
    \item [\textbf{C3.}] Lack of capability in analyzing both within-variance of groups and instances.
    \item [\textbf{C4.}] Unreliability without model quality inspection or attribute importance.
    \item [\textbf{C5.}] Less informative qualitative details.
    \item [\textbf{C6.}] Lack of supports in providing rationales on individual decisions.
\end{itemize}

These six common facets of concerns can be grouped and summarized into three main implications:
{\bf (1) A lack of bridge between top-down and bottom-up group analysis:} During the interviews, we found the discrepancy in common practices of conducting group analyses among experts. Some of our interviewees prefer to use deductive approach, e.g., using regression or other predictive modeling to test hypotheses, while others mostly use inductive approach, e.g., using clustering techniques to discover patterns not previously hypothesized. Especially for some research topics (e.g., analyzing team or demographic faultlines), clustering task tends to be a prevalent way of characterizing subgroups that are distinctive in their traits. Based on the interviews, we learned that a better tool may facilitate users not only to see the big picture (\textbf{C1}), but to efficiently trace the links between the higher-level patterns (hypothesized or not) (\textbf{C2}) and the instances located in the database (\textbf{C3}), which  
allows in-depth analysis by crossing over the current two practices of top-down and bottom-up group analysis.
{\bf (2) A lack of confidence in sophisticated tools and the analysis results:} Several interviewees expressed concerns about the lack of transparency when using sophisticated machine learning tools. For example, these techniques are not helpful for explaining the group culture with hypotheses (\textbf{C4}) or offering qualitative evidence (\textbf{C5}).
{\bf (3) A concern about the implication for group-level analyses:} A general concern repeatedly stated by the interviewees is that the correlations learned from the big data by sophisticated tools may be translated to decisions that benefit or harm certain individuals (\textbf{C6}). In particular, several of them referred to the concern of profiling under GDPR \cite{gdpr}. 

Based on the feedback from domain experts, we determine six tasks (denoted as {\bf T}) that address the aforementioned concerns, which were grouped to four guidelines (denoted as \textbf{G}) in a bottom-up manner.

\paragraph{\bf G1. Identify the shared characteristics of a group of people and the differences between groups.} The analytic tool should allow users to see how groups of people share the common characteristics and how groups differ from each other in terms of key attributes of interest.

\begin{itemize}
    \setlength{\itemsep}{0.2pt}
\item[{\bf T1}] \textbf{Group trend}: in our design, the characteristics of groups will be visualized as ``group trend'' and the differences between groups will be contrasted through visual encoding.
\item[{\bf T2}] \textbf{Inference reliance}: the tool will support users to assess the analytic model quality against ground-truth data whenever available. 
\end{itemize}

\paragraph{\bf G2. Inspect the shared characteristics and variance of groups in both quantitative and qualitative ways for hypothesis testing and searching.} The tool should allow users to closely inspect the group differences in two ways: it should provide the quantitative evidence showing how the key characteristics or attributes differentiate the groups, and the qualitative details where specific instances from the data can be retrieved to corroborate the identified characteristics. It should also allow users to discover variance from the explicit grouping to reduce the possible overgeneralization of the group distinctions.

\begin{itemize}
    \setlength{\itemsep}{-0.5pt}
\item[{\bf T3}] \textbf{Group variance}: the tool will extract and visualize subgroup characteristics to support the examination of the within-group trends and variation.
\item[{\bf T4}] \textbf{Attribute importance}: how groups are differentiable by the key characteristics or attributes will be visualized as ``attribute importance'' -- the dependence of a given attribute when predicting a group. 
\item[{\bf T5}] \textbf{Qualitative details}: the tool will enable users to retrieve the qualitative cues from individual data instances that are representative for each quantitative measured attribute.
\end{itemize}

\paragraph{\bf G3. Provide the rationale for the prediction of an instance as a group member.} The tool should allow users to understand the rationale behind each individual's group assignment produced by any analytic/predictive models. The rationale should allow users to connect to or verify the identified group characteristics. The tool should provide a module to offer instance-level explanation, which is complementary to the approaches for describing model behavior with feature importance (in \textbf{G2}) or aggregated individual explanations \cite{krause2018user, stumpf2016explanations, poursabzi2018manipulating}.

\begin{itemize}
    \setlength{\itemsep}{0.2pt}
\item[{\bf T6}] \textbf{Grouping rationale}: in our design, the group rationale will be offered through a ``contrastive explanation'' -- to explain why a member is considered to belong to one group rather than the other.
\end{itemize}

\section{Application Scenario and Data} \label{sec:appanddata}

We take the aforementioned data journalist example as an application scenario to design a visual analytic system following the design guideline. In this scenario, a user (the data journalist) is interested in exploring Twitter users' communications related to gun and gun-control issues. The user wonders: ``{\it how do social media users with different political leanings talk differently on the gun issues?}'' ``{\it how do the differences revealed through the hypothesized attributes -- particularly certain sociolinguistic characteristics -- and manifested on users' tweets?}'' and ``{\it how can I make sense of and trust the group analysis results?''}

We make use of a publicly available Twitter dataset from a prior study \cite{yan2017quantifying}. This dataset contains more than 600k Twitter users that have been identified with liberal or conservative leaning based on their following behaviors \cite{yan2017quantifying,yan2020mimicprop}. From this data, we have identified a total of 3,100 tweets and 2,256 users that are related to ``gun rights'' or ``gun control'' discussions. In addition to standard keyword matching method, we use a PU learning algorithm \cite{fusilier2015detecting} to identify related tweets, then validate the results manually. In order to verify the machine-inferred sociolinguistic characteristics from the tweet text, each tweet was manually annotated with seven sociolinguistic attributes, including \textit{Care}, \textit{Authority}, \textit{Purity}, \textit{Fairness}, and \textit{Loyalty} as described by Graham et al. \cite{graham2009liberals} as moral values, and \textit{Valence} (the happiness expressed) and \textit{Dominance} (the degree of control exerted) \cite{warriner2013norms} as affects. We summarize the definitions of these seven attributes and their annotation process in the Appendix (Section~\ref{sec:attr-def} and \ref{sec:annotation}). This carefully curated dataset allows us to design and test our new visual analytic toolkit with the set of {\it ground-truth} group labels and attribute values in a real analysis scenario. In the following sections, we use ``\red'' and ``\blue''  as the group labels.

\begin{figure}
    \vspace{-1em}
    \includegraphics[width=\columnwidth]{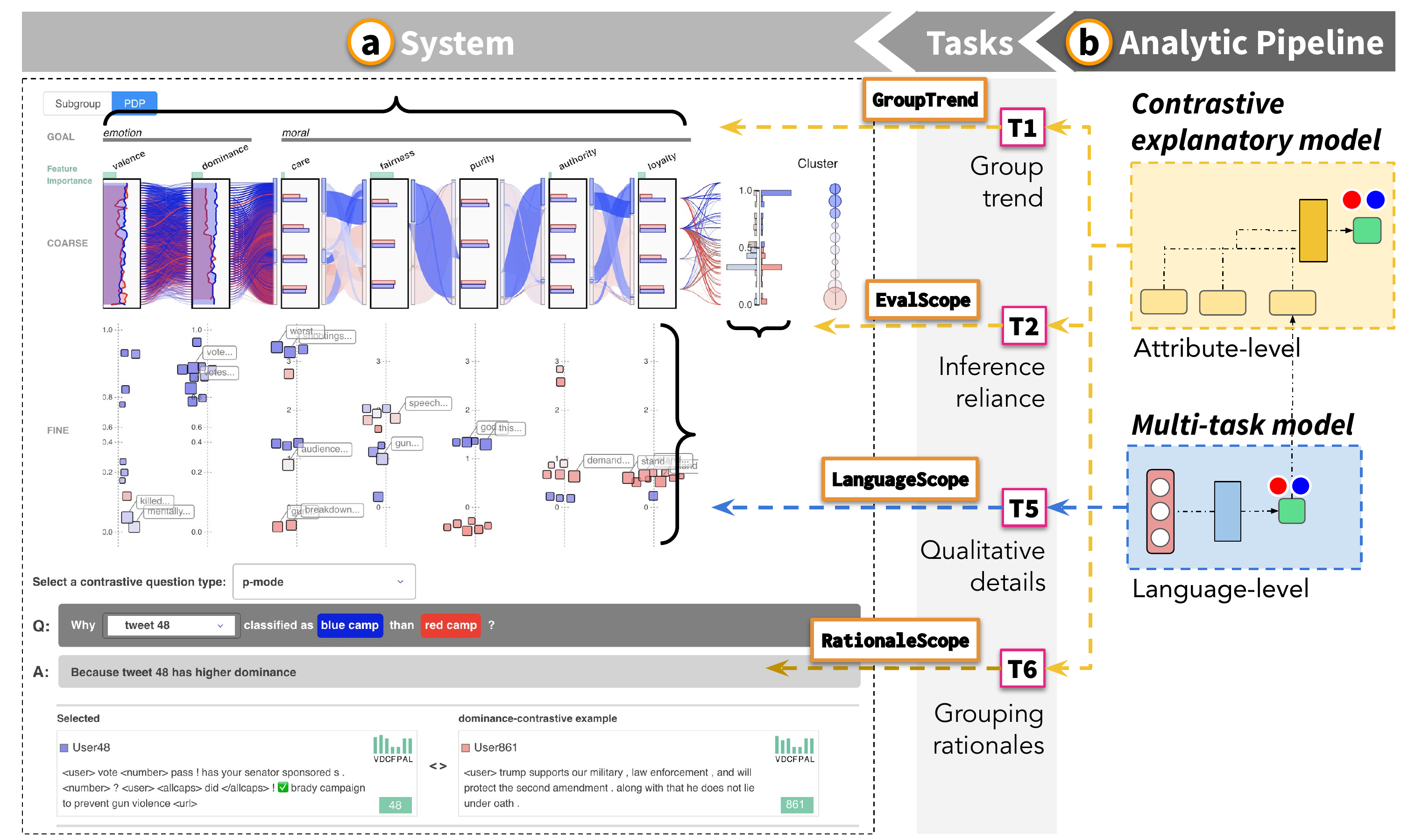}
    \vspace{-1em}
    \caption{\label{fig:system-task-analytic-pipeline}
    Our system in (a) operates on the analytic pipeline in (b) consisting of: \textit{contrastive explanatory models} for predicting and discovering the group difference being rendered in \grouptrend and \evalscope, and providing rationales for explaining the prediction of any instance in \rationalescope, and \textit{multi-task prediction models} for generating language cues from the tweet dataset to support the visual inspection in \languagescope.
    }
\end{figure}

\section{\name}
\name is built following the design guideline described earlier. As shown in Fig.~ \ref{fig:system-task-analytic-pipeline}, \name consists of two main components:  (a) visualization and (b) analytic pipeline. The visualization component is comprised of a number of visual interactive tools to support the six main tasks described in Section~\ref{sec:goalsandtasks}, including \grouptrend ({\bf T1}), \evalscope ({\bf T2}), \variscope ({\bf T3}), \depscope ({\bf T4}), \languagescope ({\bf T5}), and \rationalescope ({\bf T6}) (Fig.~\ref{fig:teaser}). Together, these six ``scopes'' allow users to access to qualitative differences ({\bf G1}) and variability ({\bf G2}) of groups, qualitative details from user-generated text ({\bf G2}), and model explanation ({\bf G3}). The analytic component (Fig. \ref{fig:system-task-analytic-pipeline}) includes two major machine learning modules: (1) {\it multi-task prediction models}  generate language cues from the tweet dataset to support the visual inspection in \languagescope, and (2) {\it contrastive explanatory models} generate rationales for explaining the membership prediction. Built up as a web-based tool, our system was implemented as a full-stack application with a python based back-end framework called Django \footnote{https://www.djangoproject.com/} to process API calls and data processing, and a front-end framework called ReactJS and React hooks\footnote{https://reactjs.org/}, and Postgres database \footnote{https://www.postgresql.org/}. Any dataset with all features and metadata is required to be stored in a file and processed into the database before running the tool. Below, we summarize our implementation of the visualization and analytic components.

\subsection{Visualization}\label{sec:vis}
Fig.~\ref{fig:system-overview} captures the user interface of \name. To facilitate users to navigate the dataset and select data/attributes of interest, two interactive tools, \instanceviewer and \scopecontroller are provided as shown on the left and the top of the user interface (Fig.~ \ref{fig:system-overview}a,b).
Once users specify the attributes of interest (e.g., which features, or sociopsychological dimensions) through the \scopecontroller, the main visualization panel with various ``scopes'' will be updated accordingly. These scopes provide distinct functionality to support tasks ({\bf T1}--{\bf T6}) in the design requirements. 
The main visualization panel seamlessly integrates multiple scopes so that the users' data exploration can be loosely guided. For example, \grouptrend can guide users' gaze from left to right through its polylines, and from top to bottom through its vertical axes. This integrated visual layout is designed to make users be contentious about what the group patterns entail. When staring at certain group patterns along the horizontal direction, users can be immediately hinted by the variability and predictive confidence of the patterns via \variscope and \evalscope. Along the vertical direction, users can easily find evidence for the patterns via \languagescope or \rationalescope.
These views also help users to make connection between the specific observations from an individual scope view and the overall group patterns. We describe the specific functions of each scope below.

Unless otherwise specified, \redclr and \blueclr colors are used to differentiate the group membership, with color saturation representing the proportion of tweets in the \red or \blue groups. 

\begin{figure*}[!ht]
    \centering
    \includegraphics[width=0.95\linewidth]{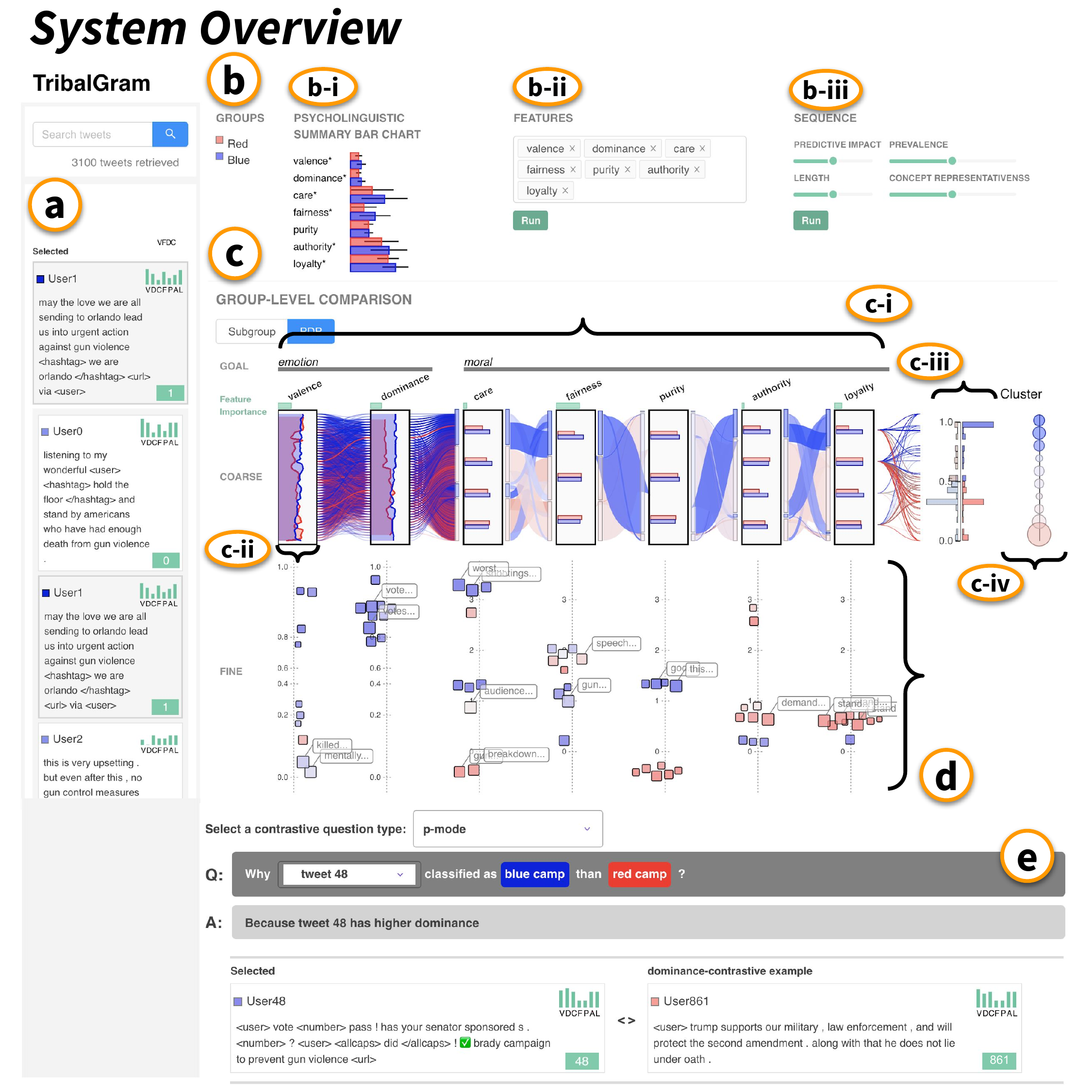}
    \caption{\label{fig:system-overview}
    The system overview of \name. The system integrates visualization and analytic pipeline to support the group analysis tasks. On the user interface, (a) \instanceviewer and (b) \scopecontroller support navigation, data retrieval, selection, and control. The data and attributes of interest are visualized in the main visualization panel through various ``scopes.'' For example, (c) \grouptrend visually captures the major ``trends'' of groups across attributes, and (d) \languagescope provides a visual summary of the language evidence for every sociolinguistic attribute, enabling users to further retrieve qualitative details from the tweet instances. (e) \rationalescope provides the instance-level explanation on why a tweet was classified as certain group in comparison with another tweet.
    }
    \label{fig:system-overview}
    \vspace{-0.8em}
\end{figure*}

\subsubsection{\bf Retrieve, Select \& Control}
Once the system is initialized, it allows users to retrieve tweets and select a particular tweet to take a closer look at it (in \instanceviewer), and control which tweets and attributes to include in the analysis (in \scopecontroller). First, \instanceviewer (Fig.~\ref{fig:system-overview}a) allows users to search and retrieve tweets by search terms. The retrieved tweets are displayed as a list of boxes. Each box contains a tweet with its group information on the top-left corner (a square glyph with its annotated group label colored as \redclr or \blueclr), the ``psycholinguistic score chart'' on the top-right corner (where each bar height represents annotated attribute value), and the tweet text. This \instanceviewer also serves as a selection tool allowing users to look into a particular tweet. For example, users can click the ``tweet handle'' located at the bottom-right corner to highlight the tweet in \grouptrend, or click an attribute on the psycholinguistic score chart to highlight the language sequence corresponding to the selected attribute in the tweet text. \scopecontroller (Fig.~\ref{fig:system-overview}b) provides an overview of all the available group attributes with statistical significance information in a ``Psycholinguistic Summary Bar Chart'' (Fig. \ref{fig:system-overview}b-i). Such overview helps users to determine which attributes to be included in (or excluded from) further group analysis. The selection of attributes can be done using the ``Features'' menu (Fig.~\ref{fig:system-overview}b-ii). The ``Sequence'' menu includes four sliders allowing users to determine the important aspects of retrieved language cues, which will be described later in the analytic modules. 

\begin{figure}
    \vspace{-1em}
    \includegraphics[width=0.95\columnwidth]{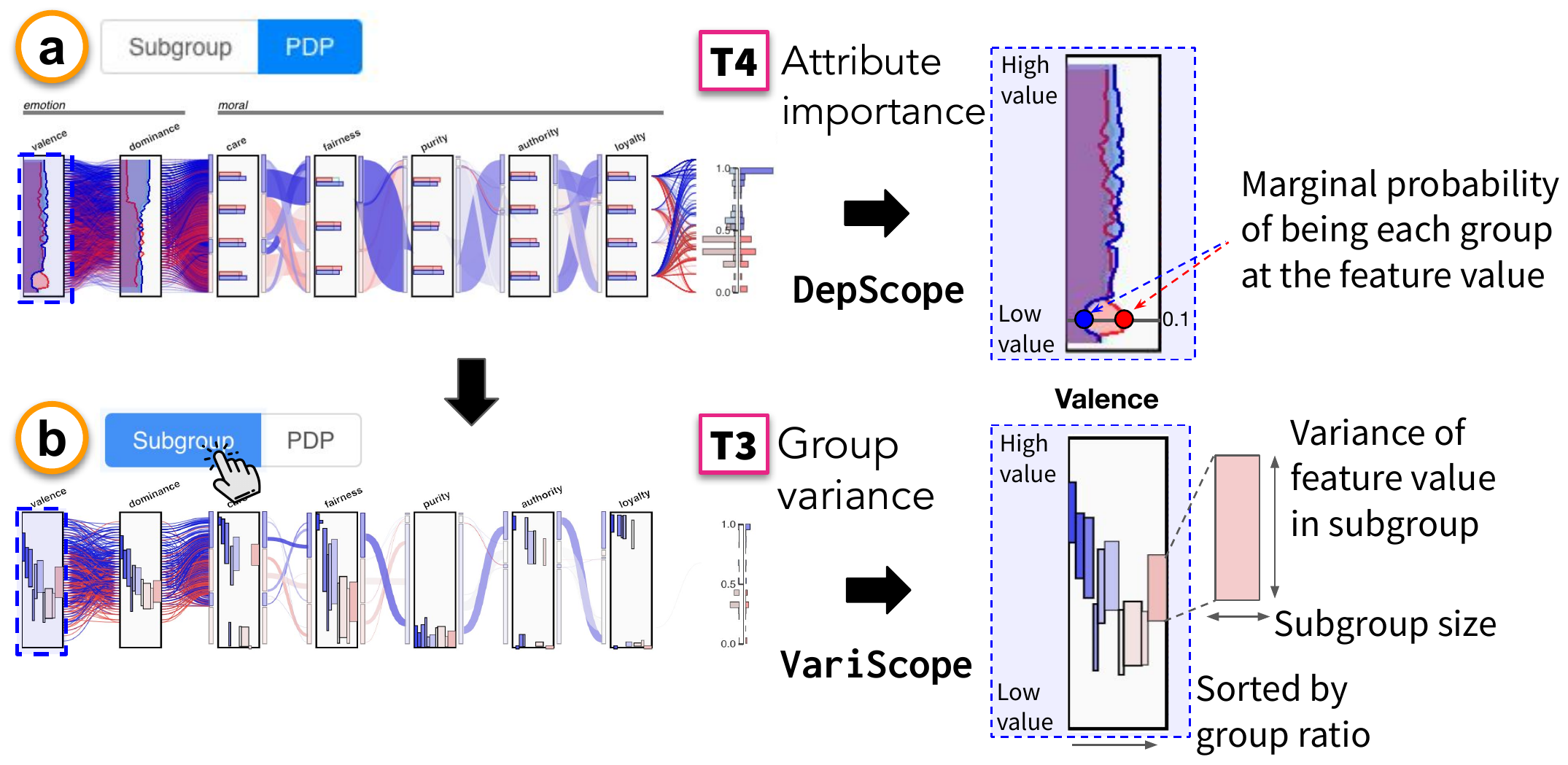}
    \vspace{-1em}
    \caption{\label{fig:system-overview-depscope-variscope}
    Two modes of Axes in \grouptrend: \depscope and \variscope.
    }
    \vspace{-1.5em}
\end{figure}

\subsubsection{\bf Overview and Inspect group difference and variance.} After setting up the configuration of analysis in \scopecontroller, users can overview the group difference in \grouptrend (Fig. \ref{fig:system-overview}c-i) with \depscope (Fig. \ref{fig:system-overview-depscope-variscope}a) or \variscope (Fig. \ref{fig:system-overview-depscope-variscope}b) as its axes in rectangular boxes allowing the inspection of feature importance and subgroups. The system also supports the inspection of the model inference results in \evalscope (Fig. \ref{fig:system-overview}c-iii).
\paragraph{\bf \grouptrend}
\grouptrend (Fig. \ref{fig:system-overview}c-i) allows users to visually capture the ``trends'' of groups and contrast the differences between them through a parallel set ({\bf T1}), where the trends are captured when polylines from a group, representing the multidimensional attribute values of data instances, agglomerate due to close attribute values. In \grouptrend, each psycholinguistic attribute is represented as a vertical axis; tweets are represented as polylines, colored by their annotated group membership, and bundled whenever appropriated to reduce the visual clutter and to enhance the rendering performance.

\paragraph{\bf \depscope \& \variscope}
The parallel axes in the \grouptrend are further augmented with \depscope and \variscope, to provide an in-context inspection of group variability on top of the group trends. Users can switch between the two modes as shown in Fig.~\ref{fig:system-overview-depscope-variscope}. The \depscope (Fig.~\ref{fig:system-overview-depscope-variscope}a) allows users to inspect the attribute importance through a ``conditional partial dependence plot'' (PDP) where the marginal probability density of each group is shown on the $x$-axes against the attribute values on the $y$-axes. This enables users to visually assess the extent to which the groups are differentiable by a given single attribute ({\bf T4}), which supports the evaluation of a hypothesis relevant to this attribute. \variscope (Fig.~\ref{fig:system-overview-depscope-variscope}b), on the other hand, allows users to inspect the group variance through a set of ``subgroup attribute glyphs'' ({\bf T3}), where each rectangle glyph represents the attribute summary of a subgroup with vertical position indicating the central tendency, height indicating the variance, width indicating the size of the subgroup, and color reflecting the probability of group membership. The subgroups were automatically detected based on the attribute values of tweets using Agglomerative Hierarchical Clustering method with the number of subgroups determined by the elbow method evaluated with the total intra-cluster variation. This enables users to visually capture the coherence or variability within a group and across attributes -- e.g., a less coherent group will have several subgroups spreading vertically along one or more attribute axes. Moreover, it serves as a hypothesis evaluation and seeking tool as the subgroup patterns may support/disconfirm an existing hypothesis, and any emerging, cross-attribute subgroup tendency may inform a new hypothesis. 

\paragraph{\bf \evalscope}
\evalscope (Fig. \ref{fig:system-overview}c-iii) allows users to closely examine the model inference results of tweet group membership by comparing the predictive membership against the ground-truth (human annotated) labels (\textbf{T2}). The predictive membership is generated from a decision tree model, which is the same as the contrastive explanatory model described later in Section~\ref{sec:contrastive-exp}. The comparison is achieved through a ``dual-sided histograms'' of classification probability (from top to bottom: from the most likely \blue to the most likely \red), where correct and wrong classifications are separately shown on the left and right side of the histogram plot. To facilitate the inspection of particular prediction cases, users can click any location of the histogram bars to highlight a particular set of instances within the corresponding range of classification probabilities.  

\subsubsection{\bf Back up with qualitative details.}
\languagescope (Fig. \ref{fig:system-overview}d) provides a visual summary of the language evidence for every sociolinguistic attribute, which enables users to capture and further retrieve qualitative details from the tweet instances (\textbf{T5}). \languagescope is comprised of multiple parallel axes arranged in the same way as the \grouptrend. The language cues are text snippets extracted from the tweet text to represent a particular attribute having a particular value (or value range) -- e.g., the text snippet \tweet{out until the house votes to address gun violence} are extracted to represent an expression for attribute \domi around the value 0.85 (Fig. \ref{fig:teaser}d). The language cues are represented as squared glyphs along each attribute axis. Each glyph is represented with its size indicating the sequence importance (described in Section~\ref{sec:multi}), with color indicating the group membership probability, and vertical position indicating the mean attribute value. These language cues are learned automatically from the multi-task prediction models, which will be described in a later section. The extracted cues are associated with an importance score that reflect both the model prediction and users' preference (as described in the \scopecontroller). By default, the system displays ten most important language cues for each attribute and the cues with an importance score greater than a threshold will be shown with the corresponding text snippets. 

\subsubsection{\bf Offer the rationale behind group profiling.}
\rationalescope (Fig. \ref{fig:system-overview}d) allows users to closely examine how individual tweets may be predicted to be in a certain group (or not) according to its attribute values, through a ``contrastive explanation dialog box'' ({\bf T6}). The dialog box presents a ``rationale'' as an answer to a question about the prediction. Users can select tweets of interest (from browsing the \instanceviewer or other scopes) and ask two types of reasoning questions \cite{van2002remote}: (1) {\it p-mode}: focuses on the ``properties'' of an object or instance (e.g., ``{\it Why is tweet X classified as \red rather than \blue}?''), and (2) {\it o-mode}: focuses on the contrast of two objects (e.g, ``{\it Why are tweet X classified as \red whereas another tweet Y classified as \blue }?''). Such rationales are automatically generated from the contrastive explanatory models, which will be described in a later section. The extracted rationales are shown using natural language with counterfactual examples in the dialog box. \\

\begin{table}[]
\resizebox{\textwidth}{!}{%
\begin{tabular}{llll}
\hline \\[-10pt]
\textbf{Layout} &
  \textbf{Design scheme} &
  \textbf{Advantage} &
  \textbf{Disadvantage} \\[2pt] \hline \\[-7pt]
\textbf{\begin{tabular}[c]{@{}l@{}}Scatterplot\\ matrix\end{tabular}} &
  \begin{tabular}[c]{@{}l@{}}A layout with a set of scatter plots \\ representing bivariate relationships \\ in a two-dimensional matrix\end{tabular} &
  \begin{tabular}[c]{@{}l@{}}Well-represents the pairwise \\ relationship between attributes\end{tabular} &
  \begin{tabular}[c]{@{}l@{}}Does not show the trends \\ over multiple attributes\end{tabular} \\[15pt]
\textbf{\begin{tabular}[c]{@{}l@{}}Radial\\ layout\end{tabular}} &
  \begin{tabular}[c]{@{}l@{}}A two-dimensional plot encoding \\ multivariate attributes of instances \\ along with radial axes\end{tabular} &
  \begin{tabular}[c]{@{}l@{}}Reveals the instance-level \\ differences by the overall \\ association with attributes\end{tabular} &
  \begin{tabular}[c]{@{}l@{}}Does not reveal \\ the attribute-wise trend\end{tabular} \\[18pt]
\textbf{\begin{tabular}[c]{@{}l@{}}Glyph-based\\ layout\end{tabular}} &
  \begin{tabular}[c]{@{}l@{}}Instances are represented as glyphs \\ encoding attribute values\\ (e.g., in case of a glyph design \\ with small radial bars)\end{tabular} &
  \begin{tabular}[c]{@{}l@{}}Visualize both instance-wise \\ characteristics and the differences \\ over multivariate attributes \\ between instances \\ in two-dimensional plot\end{tabular} &
  \begin{tabular}[c]{@{}l@{}}Hard to show attribute-wise \\ trend and group differences\end{tabular} \\[30pt]
\textbf{\begin{tabular}[c]{@{}l@{}}Parallel\\ coordinates/\\ sets\end{tabular}} &
  \begin{tabular}[c]{@{}l@{}}A axis-based layout with its polylines\\ as instances passing through multiple \\ axes for multivariate characteristics\end{tabular} &
  \begin{tabular}[c]{@{}l@{}}Provides the overall trends \\ of instances' attributes\end{tabular} &
  \begin{tabular}[c]{@{}l@{}}Less space-efficient compared \\ to other layouts\end{tabular} \\[15pt] \hline
\end{tabular}%
}
\label{table:design_choice}
\caption{\label{table:design_choice} The summary of visual layouts in the design process of \grouptrend.}
\end{table}

\subsubsection{\bf Design choice and consideration.} To decide the proper design of visual components and integrated layout methods, we went through multiple phases of the design process defining the underlying visualization tasks/problems and examining prior research especially in multi-dimensional visualization. Following the Munzner’s Nested Model \cite{munzner2009nested}, we started with defining the core domain problem, “visualizing group difference”, and the specific design requirements listed in Section \ref{sec:goalsandtasks} above. We then identified the possible data types (i.e., continuous, ordinal, or categorical attributes of group characteristics, and text data) and the operations (i.e., predictive analysis and its attribute-wise interpretation) to be considered in our system.

To support the visualization of domain problems, tasks, data types, and operations all together, we break down the whole design process into two phases: First, we define a visual layout for it to serve the core domain problem, ``visualizing the group differences''. In the system, we let \grouptrend play an central role, and investigated multi-dimensional visualization to determine the design of 
\grouptrend. Second, we define other visual components in accordance with \grouptrend to serve the aforementioned tasks and operations. In summary, in these steps we compared alternative choices of organizing visual space and layouts for multidimensional dataset and their potential to be extended to support machine learning and interpretability functionality. In the final layout, other layout components such as \depscope, \variscope, \evalscope are tightly coupled and integrated with the design of \grouptrend so that the visual layout as a whole not only serves the requirements of our system but visually associates with each other. We illustrate the two considerations of the design process in detail below.

First, to find out the design of \grouptrend for visualizing the group differences, we investigated 10 designs in the multi-dimensional visualization studies. In the literature review, we first categorized visual layouts in four types of multi-dimensional visualization based on the classifications in two surveys \cite{liuVisualizingHighDimensionalData2017a, hoffman2002survey}, then searched for literature by the name of layout types and selected 10 visual layouts. The design alternatives as a result of this process can be categorized into four typical multidimensional visual layouts: parallel coordinates and sets \cite{kosara2006parallel, richerEnablingHierarchicalExploration, vosoughParallelHierarchiesVisualization2018, weideleAutoAIVizOpeningBlackbox2020, novotnyOutlierPreservingFocusContext2006} and radial layout \cite{albuquerque2010improving, wang2019polarviz}, scatterplot matrix \cite{wilkinson2006high}, and glyph-based layout \cite{zhaoSkyLensVisualAnalysis2018, cao2018z}. 

Table \ref{table:design_choice} provides a summary of advantages and disadvantages of visual layouts examined in the design decision. Among them, scatterplot matrix is a well-known visual layout with a set of scatter plots representing bivariate relationships in a two-dimensional matrix. While it effectively reveals how data is correlated with respect to any combinations of two variables, we find that it has a limitation of showing trends throughout multiple variables in our application. Radial and glyph-based layout are other types of layout which encode multivariate attributes of an instance as a vector with a point or glyph being projected and coordinated in two dimensional plot. It is advantageous by its two-sided strategy, encoding the overall similarity between instances by their coordinates and attribute-wise properties of instances by radial axes or glyphs with small radial bars, however, it does not facilitate the overview of agglomerative group-wise trends.

We compare the pros and cons of all possible visual layout candidates, as summarized in Table \ref{table:design_choice}, and decided that the parallel sets/coordinates are most suitable to meet the required data types and operations.
It provides the overview of multivariate group trends especially with polylines colored by group memberships along with multiple axes. To support heterogeneous data types, we combine the layout of parallel sets and coordinates. For example, in our dataset, the {\it affect} attributes (e.g., \vale) are continuous variables and the {\it moral} attributes (e.g., \care, \fair) are categorical/ordinal variables. When different types of variables are involved, a polyline indicates either an instance (for continuous variable) or a group of instances belonging to a category (for categorical/ordinal variable). These polylines pass through the vertical parallel axes that are arranged in a way that the highest to lowest possible values of continuous variables are shown from the top to the bottom, while the values in categorical variables are shown in the same or similar orders (e.g., ``virtue,'' ``both,'' ``none,'' and ``vice''). 

Second, considering the visualization of classification results and attribute-wise interpretation, we find that the choice of the parallel sets as a visual layout benefits from its extendability, with which we turned it into “parallel sets for classification” with several visual components being integrated and connected to \grouptrend as an extension of traditional parallel sets. Specifically, we extend the visual space of parallel sets to incorporate the classification results and interpretation in two ways: 1) Utilizing unused visual space - by utilizing the axes of parallel sets for attribute-wise interpretation (presented as \depscope or \variscope). The vertical parallel axes in the traditional layout are typically of no use without any functionality. We expand it to a vertical space to encode the attribute-wise interpretation (details in the \depscope or \variscope section), 2) Connecting to other layouts - with the histogram plot of prediction results (\evalscope) being aligned and connected on the right side of \grouptrend (as shown in Fig. \ref{fig:system-overview}c-i and Fig. \ref{fig:system-overview}c-iii). The polylines in \grouptrend flow through the vertical axes and lead to \evalscope, which can show how group trends are associated with the inference results (details in the \evalscope section).

\subsection{Analytic pipeline}\label{sec:analytic}

The analytic pipeline, as shown in Fig.~ \ref{fig:system-task-analytic-pipeline}b, includes data processing and machine learning modules to extract information to be shown on the visualization interface. As complete details of the implementation are beyond the scope of this paper, here we provide our methodology for implementing the two major machine learning modules. We propose two machine learning algorithms to enhance the interpretability and explainability of group analysis: (1) Multi-task predictive model: We introduce a multi-task prediction neural architecture predicting jointly both group membership and attribute values from language sequences enables to extract attribute-wise linguistic cues as qualitative evidence from the attention mechanism with better predictive performance. (2) Contrastive explanatory model: We provide the module of generating contrastive explanations to present the minimal and sufficient information of group classification results. By leveraging a contrastive explanation approach \cite{van2018contrastive}, our pipeline introduces our criteria and methods to retrieve counterfactual examples in fact-foil tree in addition to explanation itself for better explainability.

\subsubsection{Generating Language Cues via Multi-task Prediction}\label{sec:multi}
\begin{figure}[!ht]
    \centering
    \includegraphics[width=0.95\columnwidth]{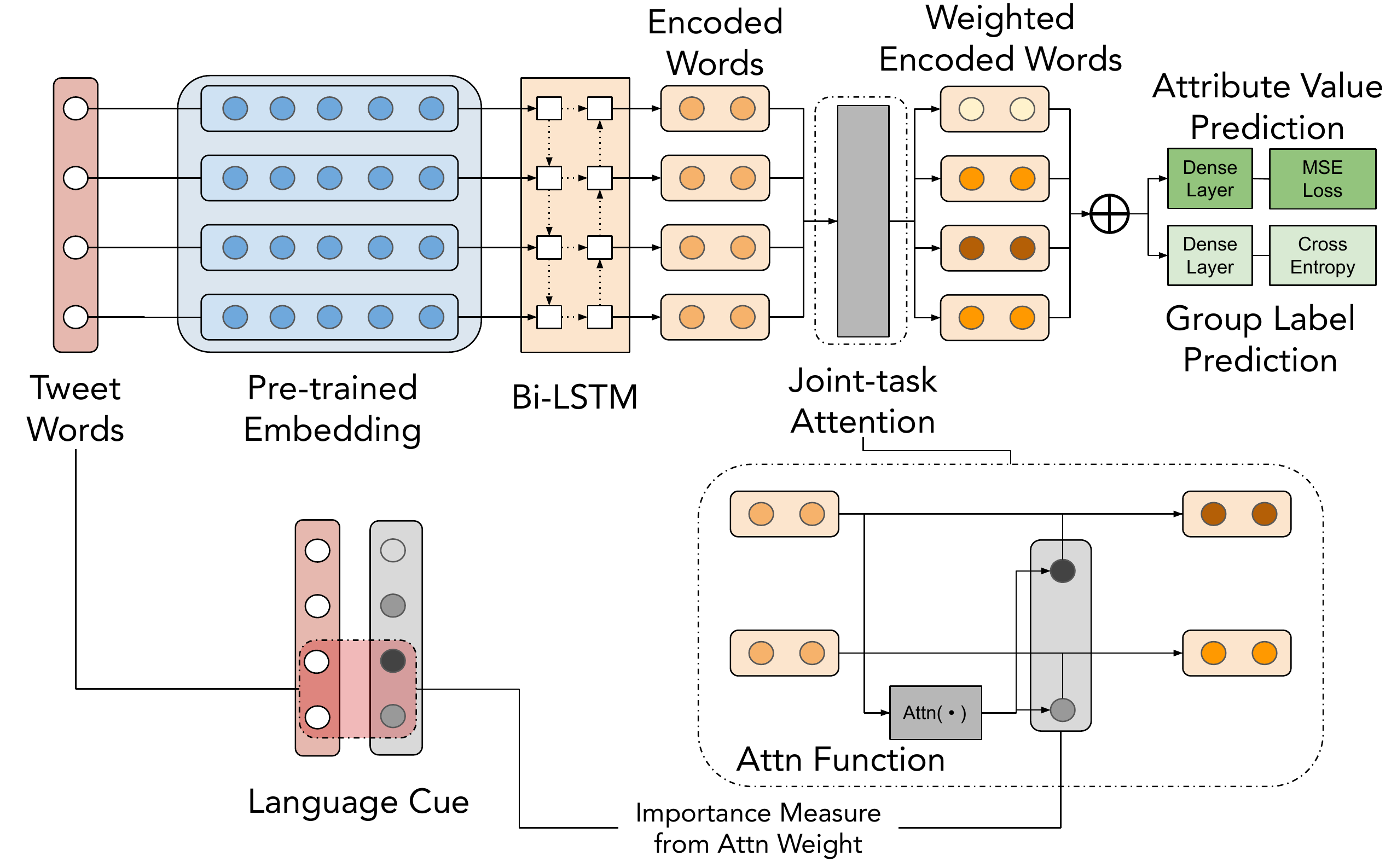}
    \caption{
    The neural network architecture for generating language cues. An input of word sequence in a tweet is transformed to word embedding and encoded by a Bi-LSTM layer, with predictive weights learned through an attention mechanism. Weighted latent vectors are taken into the dense layers to jointly predict the group and attribute value. Informative language cues are generated from a function of the learned attention weights. 
    }
    \label{fig:multi-model}
\end{figure}
The {\it multi-task prediction models\footnote{The code is available at: \url{https://github.com/picsolab/TRIBAL-multi-task-prediction}}} are developed to extract text snippets as language cues from the tweet dataset. Compared to language models with a bottom-up approach for discovering latent dimensions of semantics such as topic modeling, our model was designed to enable the theory-driven analysis where attributes to be included in the analysis are given by users in a top-down manner, and allow them to find linguistic cues with respect to each attribute. Our model is thus advantageous in interpreting the group difference based on (a) attributes that are  theoretically meaningful (e.g., emotional/moral attributes) or (b) data-driven features indicative of behavioral patterns (e.g., the number of retweets). Language cues identified from the model allows users to make sense of an attribute (e.g., ``{\it What dose high \vale mean in this context}?'') through retrieving qualitative details from the tweet instances (e.g., ``{\it What would the language expressing high \vale look like from the \red or \blue groups}?'')  ({\bf T5}), which are shown in the \languagescope. To automatically learn the language cues that are {\it representative for both groups and attribute values}, we introduce a new multi-task prediction neural network architecture, where the objective is to jointly predict (a) attribute values and (b) group labels (\red or \blue) from the language sequence of a given tweet.  As a result, the neural network architecture can learn to identify what language sequences are more predictive to a particular group and attribute information. Because the representative language cues for different attributes will be different, we train a set of models with similar architectures but different objective functions (one model for each attribute).

Take the \vale attribute as an example. Fig~\ref{fig:multi-model} illustrates the neural network architecture that jointly predicts \vale value and group label. The input tweet text is represented by a pre-trained embedding, where each tweet word is represented as a high-dimensional vector. Due to the specific emphasis on sociolinguistic values in this context, we leverage two pre-trained embedding methods: (1) a word2vec \cite{mikolov2013efficient} embedding trained on a standard Twitter corpus \cite{baziotis2018ntua}, and (2) the attribute-aligned embedding trained with MimicProp algorithm \cite{yan2020mimicprop} that is optimized for sociolinguistic lexicons. We concatenate the two equal-sized embeddings to generate a 600-dimensional vector for each word.

Let the $\mat{X}= \{\vec{x}_{1}, \vec{x}_2, …, \vec{x}_t, ... \vec{x}_n\}$ represent the embedding language sequence for a tweet with $n$ words,  where $\vec{x}_t$ represents the embedding vector for the $t^{th}$ word in the tweet. Let $l$ and $s$ denote the ground-truth label and attribute value of a tweet. The objective is to minimize the total loss per tweet:
\begin{equation}
    loss = \lambda\cdot loss_{group}  + (1 - \lambda)\cdot loss_{attr},
\end{equation}

where the $loss_{group} = CrossEntropy(pr_{group}, l)$ is the cross-entropy loss for predicting group label by comparing the posterior label probability $pr_{group}$ with the true label $l$, and $loss_{attr} = MSE(logit_{attr}, s)$ is the mean squared error for predicting attribute value by comparing the logit $logit_{attr}$ with the true value $s$, and $\lambda$ is a hyper-parameter determining the trade-off between two types of loss. 

For continuous attributes (e.g., \vale in this case), we use $MSE(\cdot)$ to model the loss, whereas for categorical attributes, the loss is computed using $CrossEntropy(\cdot)$ similar to the loss for group prediction. We leverage a bi-LSTM encoder \cite{huang2015bidirectional} with an attention mechanism \cite{cho2014learning} in order to learn {\it which specific portion} of the language sequence serves predictive features to the prediction. The bi-LSTM encoder learns a latent representation recursively for an input word at location $t$, as $\vec{h}_t = f[(\vec{x}_t) , \vec{h}_{t-1}]$, with $\vec{h}_{0}$ a trainable bias parameter. The latent representation is then taken as the input for the attention layer to learn the attention weight $a_{t}$ by: $a_{t} = \frac{exp(e_{t})}{\sum_{k = 1}^{n}exp(e_{k})}, e_{t} = attn(\vec{h}_{t})$, where $attn(\cdot)$ is a dense layer with trainable weights $\vec{w}$ and bias parameter $b$ which transforms $\vec{h}_{t}$ as $e_{t} = \vec{w} \cdot \vec{h}_{t} + b$.
The latent vector $\vec{h}_{t}$ is then weighted by the learned attention weight $a_t$ as an input for the dense layers for computing the loss. 

The attention weights can be seen as the importance of each word in the prediction. As our goal is to retrieve a sequence (of consecutive words) rather than individual words, we compute the cumulative attention for a length-$k$ sequence $\vec{w} = \{w_i … w_{i+k-1}\}$ with attention weights $\vec{a} = \{a_i ... a_{i+k-1}\}$ where the length is determined when adding the next word to the current sequence decreases the mean attention of the sequence to a great extent (specifically, $k$ is automatically detected when increase of $k$ leads to the drop of mean attention exceeding a threshold $\theta = a_t - \frac{1}{2} \sigma^2(\vec{a})$). 

We evaluate our multi-task prediction models using a hold-out experiment. The experiment results suggest that our multi-task models can significantly improve group prediction without sacrificing the performance for single attribute prediction. 
In the experiments, we determine the hyperparameter $\lambda$ by searching on the space from 0 to 1 with 0.1 interval.
Our empirical results suggest 0.5 to be the optimal $\lambda$.
Experiment results reported in Section~\ref{sec:holdout} are all from models with $\lambda=0.5$.

Lastly, to compare the ``representativeness'' among all language sequences extracted from the data through this process, we define an importance score for each sequence based on four information criteria: (1) {\it Predictive Impact} reflects the predictive contribution of the sequence across all training tweet samples, which is computed as the percentile rank of the aforementioned cumulative attention for an extracted sequence. (2) {\it Concept Representativeness} captures how closely the tweet containing the sequence is aligned with the joint objective, which is measured using the normalized posterior probability of the tweet. (3) {\it Length} is the desired length of the extracted sequences, where longer sequences are usually preferred. (4) {\it Prevalence} reflect the frequency of an extracted sequence. While similar or identical sequences in the training data may aggregately achieve higher predictive power, retrieving similar sequences adds little to human interpretation. We thus consider the relative occurrence of a sequence in the data as an indicator for its (lack of) uniqueness. Together the four criteria are used to compute the sequence importance score as:
$\vec{w}_{seq} = \frac{p + log(l + \epsilon)}{r + log(f + \epsilon)}$,
where $\epsilon$ is a smoothing term, $p, l, r$ and $f$ denote the \textit{\textbf{p}osterior score}, \textit{sequence \textbf{l}ength}, \textit{attention weight \textbf{r}anking} and \textit{sequence \textbf{f}requency}, respectively. To enable users to retrieve language cues with different characteristics, the weights of the four criteria can be adjusted in the \scopecontroller as described in Section~\ref{sec:vis} (Fig.~\ref{fig:system-overview}b-iii).

\subsubsection{Generating Rationales via Contrastive Explanatory Models}\label{sec:contrastive-exp}

The {\it contrastive explanatory models} was developed to generate a rationale behind the membership classification of any given tweet ({\bf T6}). Instead of offering a more complete and well-rounded explanation, in this work, we leverage the contrastive explanation approach proposed by van der Waa et al. \cite{van2018contrastive} to generate a simple interpretation in a more user-friendly manner -- to present the minimal and sufficient information required to understand the current output by contrasting with another one that is absent. Specifically, we use fact-foil trees (locally trained one-versus-all decision trees) to identify the disjoint set of rules on important features to answer a question like ``{\it why this output (the fact) instead of that output (the foil)}?'' We further contextualize such question in our application scenario to produce two types of questions -- the {\it p-mode} (``{\it group \red instead of group \blue}?'') and {\it o-mode} (``{\it tweet X instead of tweet Y}?'') questions as described in \rationalescope (Section~\ref{sec:vis}). Our \textit{contrastive explanatory model} consists of three steps. The first and second steps were to generate a set of contrastive explanations from the decision rules based on the fact-foil tree model \cite{van2018contrastive}. However, there could be numerous instances falling into this explanation set, and showing all the instances will be overwhelming. Therefore, we introduce a third step to select contrastive examples that best represent a contrastive explanation.
\begin{enumerate}
\vspace{-0.2em}
\item {\it Identifying a foil leaf}: In a fact-foil decision tree, the leaves are considered as a contrasting unit of classification. 
We first identify a leaf that includes the selected instance of our interest (e.g., tweet $X$ classified as \red), and then find the counterfactual leaf to contrast the selected leaf (e.g., tweet $Y$ classified as \blue), with their differences extracted as an explanation. If a tweet $X$ is placed in leaf $l$, the contrastive leaf is the closest one but classified as \blue. 
\item {\it Generate contrastive explanation with decision rules}: In the decision tree, the decision rules along the path from the root to a selected leaf can be considered as a full explanation on why a tweet was classified as a certain class. To generate a minimal and sufficient explanation, we extract only the difference between the two paths to the fact and foil leaves. 
\item {\it Select a contrastive example}: To generate a contrastive explanation to show in the \rationalescope, we need not only the minimal information from the decision rules (about what attribute or attributes are most important) but also a tweet example to illustrate such decision rules. We identify the most appropriate tweet example within the set of instances that belong to the foil leaves based on two criteria -- the closest (base on Gower distance between the tweets' attribute values \cite{gower1971general}) and reliable (correctly classified as in another class/group) tweet. 
\end{enumerate}

\section{Use Case Scenario}\label{sec:case}
\begin{figure}[!t]
    \centering
    \vspace{-0.9em}
    \includegraphics[width=0.95\columnwidth]{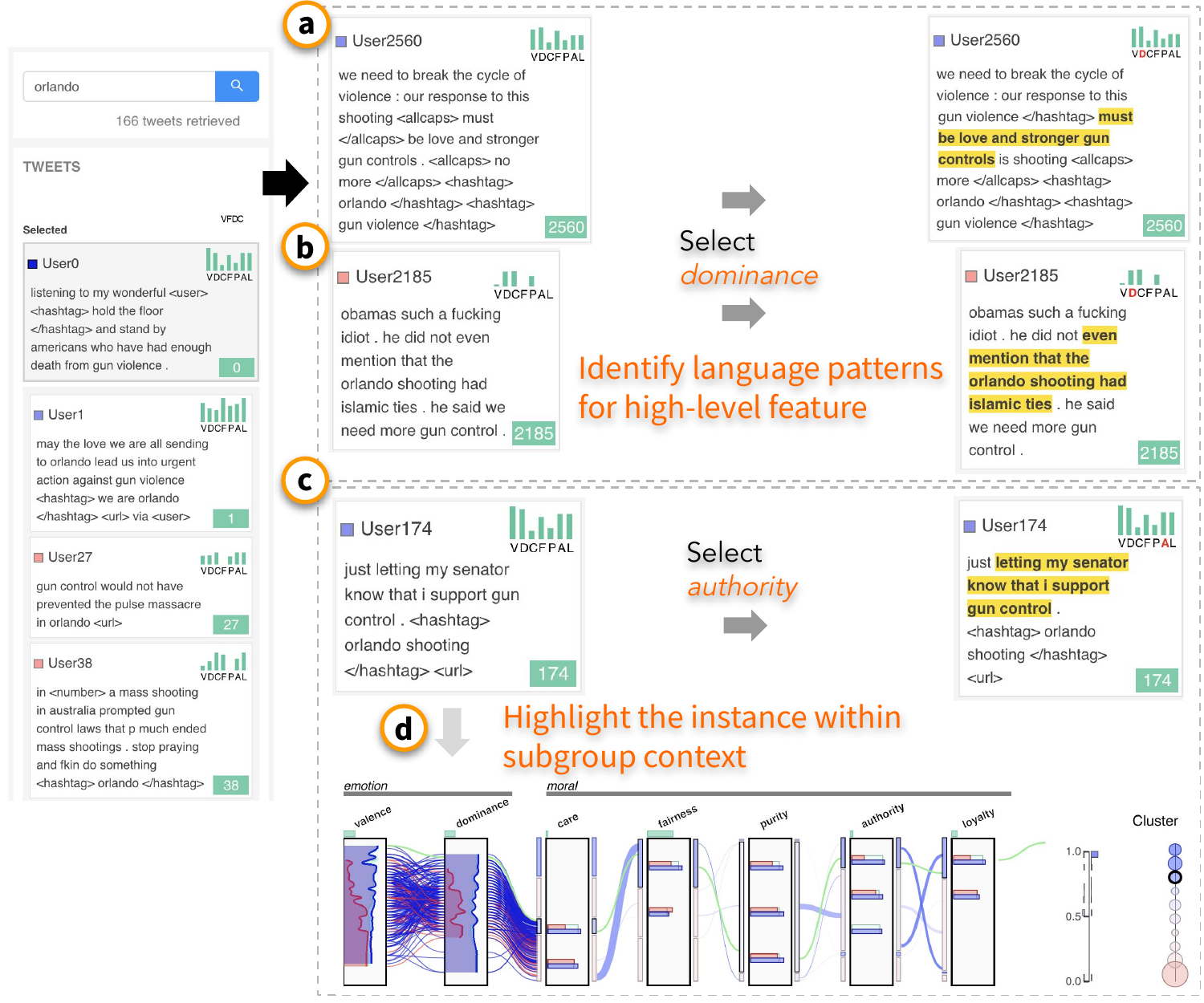}
    \vspace{-0.9em}
    \caption{\label{fig:case-study-individual}
    \textbf{Use case scenario (T5)}: While browsing tweets related to the ``Orlando shooting'' event, (a)-(b) the user explores two tweets from each group with different \domi values, and identified the expression of \domi from the psycholinguistic score chart. The user (c) explores the expression of \auth, and (d) can further compare the tweet against others within the subgroup 3 with its polyline highlighted as green in \grouptrend.}
    \vspace{-1.3em}
\end{figure}

We now present a use case scenario for how \name facilitates accountable group-level analyses. Consider the earlier data journalist example -- Erin is trying to examine the public's sentiments on social media after a mass shooting event. In particular, she would like to determine whether online users with liberal or conservative-leaning might talk differently over the topics of gun violence, gun policies, and related issues. She hopes a deep dive into online users' conversation would give her more insights in addition to the polls that have been reported elsewhere. However, merely searching tweets through a search engine does little for her goals. In this scenario, we use aforementioned Twitter data (see Section \ref{sec:appanddata} for the details of the dataset).

\paragraph{\bf Retrieve relevant tweets and qualitative cues (T5)}
Fig. \ref{fig:case-study-individual}a-b shows how she can use \name to quickly identify relevant tweets with diverse expressions. She entered ``orlando'' as a search term, expecting to find tweets about the Orlando shooting incidents happening in 2016. This event has provoked intense social media reactions, nationwide debates, and subsequent legislative actions. As a result of search, the keyword query returned 166 tweets shown on the \instanceviewer, from both the \blue and \red camps (e.g., Fig. \ref{fig:case-study-individual}a and b, respectively). To look at how these users talked about the event differently, she explored the {\it psycholinguistic summary} bar chart as shown at the top-right corner of each tweet. She found the first two tweets seemed to be quite different in terms of expressing \domi (a sense of feeling in control or losing control of a certain state). This prompted her to look for more tweets that express \domi: ``Can I find such expressions from both groups to compare?'' She clicked the bar `D' (abbr. for \domi) from those tweets (Fig. \ref{fig:case-study-individual}a), the first tweet (from User2560 in the \blue camp) had a highlighted language \tweet{must be love and stronger gun controls}, and the second (from User2185 in the \red camp) had \tweet{even mention that the Orlando shooting had islamic ties} (Fig. \ref{fig:case-study-individual}b). These illustrate how the two users had expressed the sense of in control or losing control differently when commenting on the Orlando shootings. Out of curiosity, she clicked `A' (abbr. for \auth) to see what an expression of \auth may look like (Fig. \ref{fig:case-study-individual}c) -- e.g., a retrieved tweet with highlighted text  \tweet{letting my senator know that I support gun control} suggests the user called his/her senator (authority) to take the leadership. 


\begin{figure}
  \centering
  \vspace{-0.8em}
  \includegraphics[width=0.95\columnwidth]{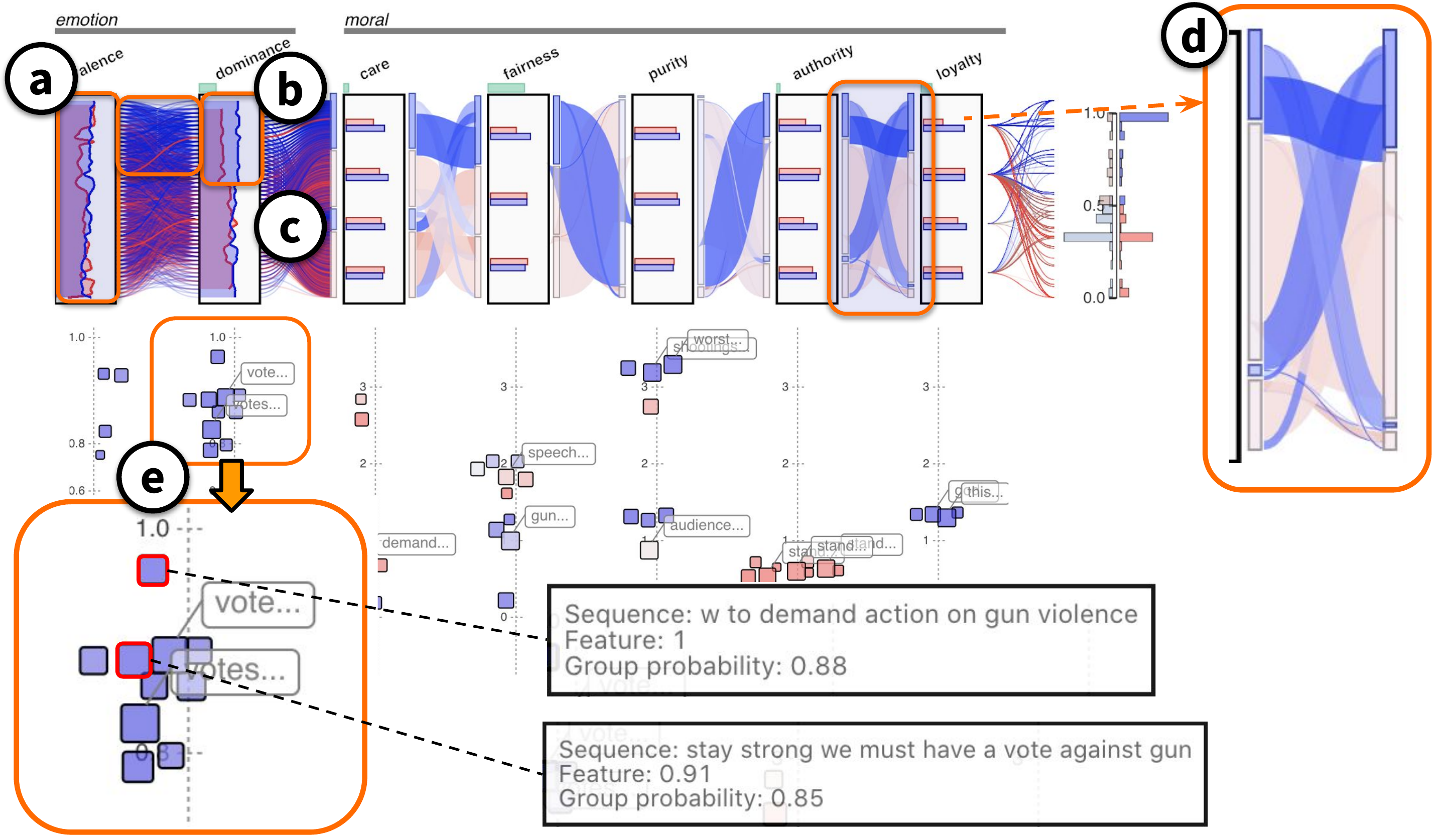}
  \vspace{-1em}
    \caption{\label{fig:case-study-group}
    \textbf{Use case scenario (T1, T4, T5)}: The user (a)-(d) explores the major trends in \grouptrend and the group-wide attribute importance in \depscope, and (e) identifies the language cues in \languagescope.}
    \label{fig:case-study-group}
\end{figure}

\paragraph{\bf Overview group trend (T1), check attribute importance (T4), and retrieve qualitative evidence (T5)}
Erin used \grouptrend to get an overview of the major differences between groups. 
In addition to the \blue camp's general associations with a higher value of \vale and \domi (Fig. \ref{fig:case-study-group}c), she found that the two camps seemed to have mixed scales in \auth and \loya, as shown by crossed edges (Fig. \ref{fig:case-study-group}d). With such observations, she now wondered whether she should focus on the two more distinguishing dimensions to determine if the two political camps had talked about the event with distinct \vale and \domi tones. To find it out, she used the \depscope to check the two dimensions separately. The partial dependence plot (PDP) in the \depscope indicated that, for \vale, a lower value tends to be associated with \red lines (Fig. \ref{fig:case-study-group}a), whereas a higher value may be mixed. Counter to her expectation, the \depscope suggested that this single dimension would not be sufficient to distinguish the two camps, as the \red appeared to have diverse values and even extreme values on both positive and negative side of \vale. On the other hand, the \domi dimension shown in \depscope was more consistent, where the \blue appeared to be associated with a higher value of \domi (Fig. \ref{fig:case-study-group}b). She became interested in telling a story about this collective tendency she observed from the \blue camp. ``{\it How can I tell the story?}'' The \languagescope allows her to track the texts in tweets with specific sociolinguistic tones. Using \languagescope, she found that the word ``vote'' was recurrently shown in the tweets with a higher \domi from the \blue camp (Fig.~\ref{fig:case-study-group}e) -- e.g., \tweet{stay strong we must have a vote against gun}. Such evidence allows her to come out with a story about how liberal-leaning users made a call to action in response to this mass shooting incident.
 
\begin{figure}[!t]
    \vspace{-1.5em}
    \setlength\tabcolsep{2pt} 
    \begin{tabular}{c}\\
    \includegraphics[width=0.95\linewidth]{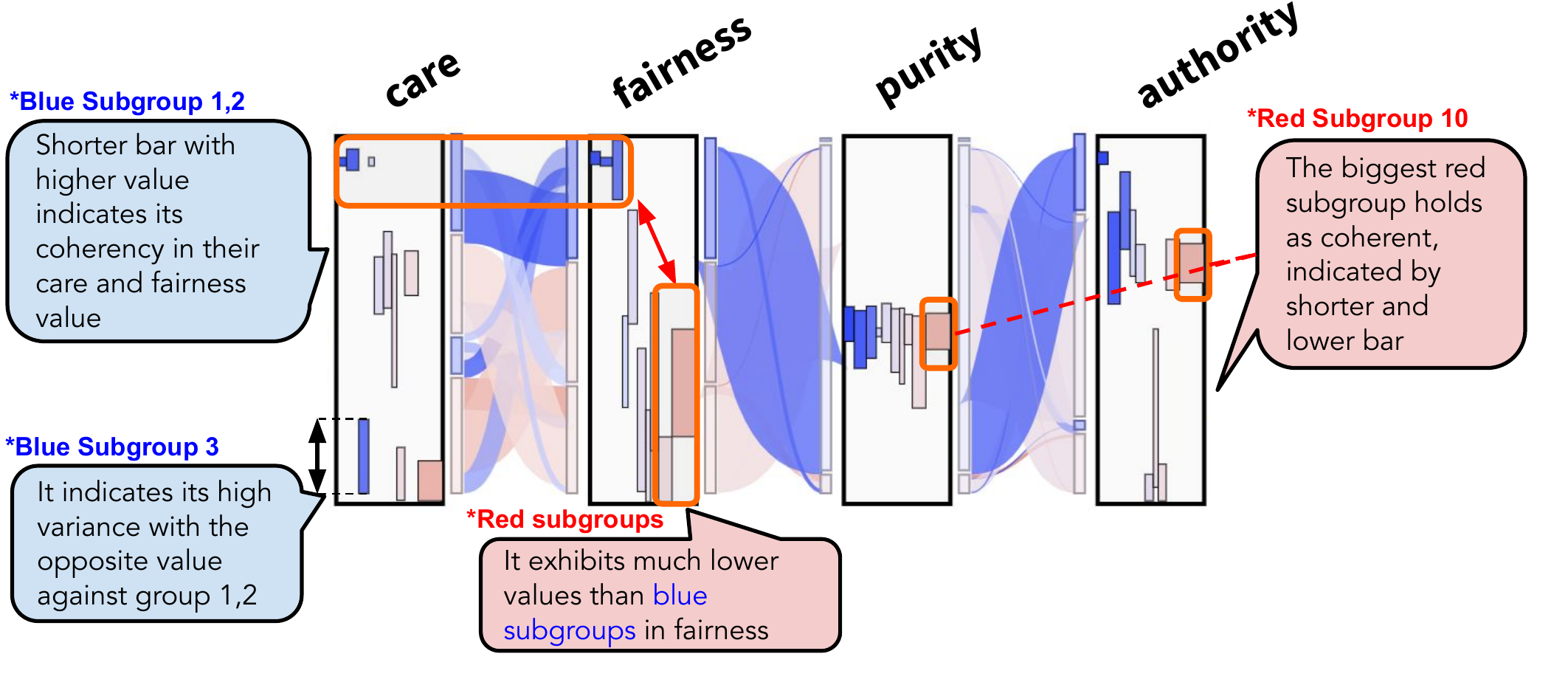}
    \end{tabular}  
    \vspace{-1em}
    \caption{\label{fig:case-study-subgroup}
    \textbf{Use case scenario (T3)}: Exploring the variance of subgroup differences using \variscope in \grouptrend.}
    \vspace{-1.5em}
\end{figure}

\paragraph{\bf Inspect group variation (T3)}
She now wondered if her story applied to all users from the \blue camp. The \variscope mode in the \grouptrend allows her to see the trends of subgroups across different sociolinguistic dimensions (Fig. \ref{fig:case-study-subgroup}). Using \variscope, she can see how subgroups (each as a rectangle bar) within the two camps may possess a higher or lower value in a particular dimension (indicated by the vertical position of the bar), and how each subgroup may have a more or less diverse pattern (indicated by the bar height) and a varying group size (indicated by the bar width). For example, she found \blue subgroups (1, 2) to have shorter bars in the \care and \fair dimensions, while the \blue subgroup 3 had a taller bar, indicating the latter subgroup had expressed different and varying tones from the rest of the \blue camp in terms of the two dimensions. Such subgroup differences prompted her to look for the subset of users who possess very similar characteristics. For the \blue camp, she found subgroup 1 seemed to be very consistent with bar positions far from other \red subgroups. For the \red camp, she found the largest \red subgroup 10 appeared to be a coherent set, located at the lower end of many features from all other \blue subgroups. 

This scenario demonstrates the major features of \name. More features will be covered in the later sections.

\section{Expert Interview}\label{sec:expert}

We conducted expert interviews to better understand whether the proposed system achieves its design goals, as well as its strengths and limitations. 
Based on the feedback from the pilot study where domain experts expressed their concerns in using existing tools that are limited by basic or surface-level group analyses, we see the evaluation process needs to be formulated in a way that demonstrates how users can gain the insights that are more complex (i.e., involving several pieces of data as evidence \textit{"in a synergistic way"} rather than simple individual data), relevant (i.e., \textit{"deeply embedded"} in the relevant domain), and deep (i.e., \textit{“accumulating and building on itself”}) while using our system. It is referred as the insight-based evaluation as termed from prior research in evaluating visualization \cite{northMeasuringVisualizationInsight2006, plaisantPromotingInsightBasedEvaluation2008}. For this purpose, instead of a quantitative evaluation we chose to conduct a more elaborate semi-structured interview to let the interviewees facilitate their thinking process enough to derive the context-specific and insightful findings simulating their workflows where they can test their own the hypothesis and find out quantitative evidence.

We invited three domain experts -- a political scientist, a social psychologist, and a machine learning expert specialized in natural language processing. All three experts had experience in working with social media data. Two of them had participated in our pilot interviews and their concerns and desired analytic support have been incorporated into our design guideline. In these open-ended interviews, we aimed to evaluate \name in a realistic group analysis workflow.

Each interview lasted about 90 minutes. The first was conducted in person, while the other two were via video conferencing. The system was running on a Chrome browser from both computers of the interviewer and interviewee. For each interview, we first provided a guided tutorial of the system and dataset, followed by a walkthrough of the system and a semi-structured interview. To emulate a realistic workflow, we asked the participants to think aloud. They were asked to consider: (1) a research question they would like to explore, or any hypothesis they may want to test or generate with the system, (2) how the system may facilitate the exploration of their question, and (3) the limitation or desired activities of current system. This section summarizes our findings from the three interviews.

\begin{figure}[!t]
    \vspace{-1.5em}
    \setlength\tabcolsep{2pt} 
    \begin{tabular}{c}\\
    \includegraphics[width=0.95\linewidth]{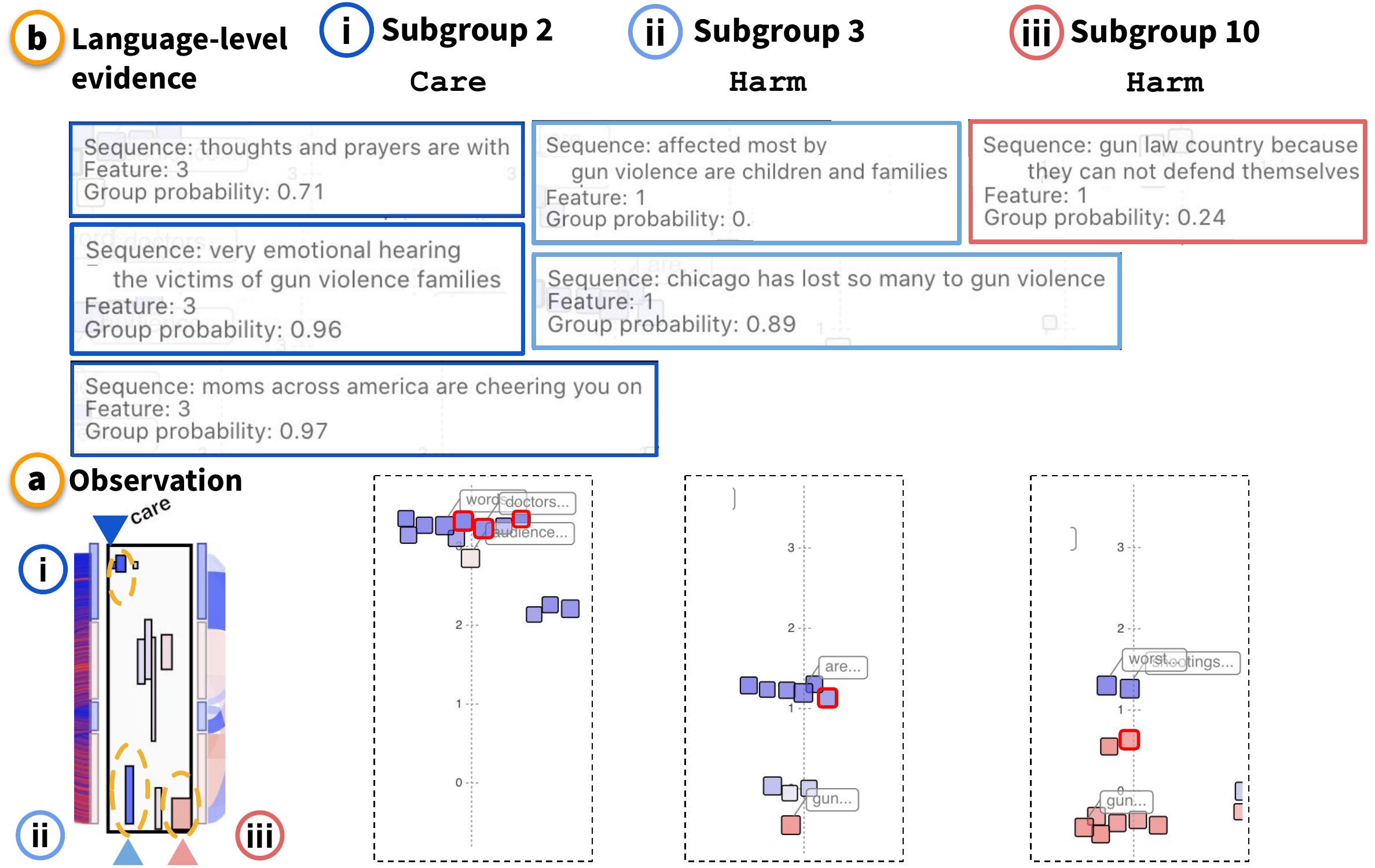}
    \end{tabular}  
    \vspace{-1em}
    \caption{\label{fig:expert-interview-1}
    \textbf{Expert interview 1}: Analyzing the within-group language variability of \care. (a) After observing the variance within three subgroups in \variscope, the expert retrieved the relevant language-level evidence to explore the aspect of issue polarization in \languagescope.}
    \vspace{-1em}
\end{figure}

\subsection{Interview 1: Exploring ways of political polarization}
Expert 1 is a political scientist interested in studying {\it the varying aspects of political polarization}. He would like to use \name to capture how social media users are polarized on gun-related issues. In particular, he wanted to explore whether the increasingly polarized online space is more of a reflection of {\it issue polarization} or {\it non-issue polarization}. He explained that in non-issue polarization, such as {\it affective} or {\it identity} polarization, the divide is driven by ideology, partisanship, or group identity, whereas in issue polarization, the group difference reflects different issue positions or policy attitudes. He hypothesized that in the case of non-issue polarization, {\it the language patterns will be more similar in one camp but mutually disjoint between camps; in contrast, users' languages will be more diverse in general if the concern is issue based}.

\paragraph{\bf Identify typical group behaviors (T1) and representative subgroups (T3)}
He started with the \scopecontroller and found the two camps were largely dissimilar in most of the sociolinguistic attributes. While confirming that the attributes of \blue were statistically significant from the \red in most of the dimensions (except for \puri) (Fig. \ref{fig:system-overview}b-i), he commented that the clear overall differences could be a sign for non-issue polarization. Next, the \variscope on \grouptrend caught his attention, ``{\it [it allows me to] take a closer look at each attribute and observe the subtle differences in each group.}'' He observed that the overall subgroup trends showed how the two groups were separated, and that the subgroups, 1 and 10, were quite ``representative'' of each camp, which represented how the attribute values of one camp were far from the other side. Having seen the varying patterns, he commented ``{\it [this could be] a useful tool for observing the partisan divide not just from political pundits but also from normal citizens.}''

\paragraph{\bf Refine initial hypotheses with language cues (T5) and help mitigate overgeneralization on group characterization}
Continuing on his exploration, he found that he learned more about ``{\it group variation rather than group coherence}'' in \variscope. For example, two \blue-dominant subgroups (2 and 3) had quite different values in the \care dimension (Fig. \ref{fig:expert-interview-1}i,ii), while \red-dominant subgroup (10) was likely to express with a tone contrary to \care (i.e., {\it harm}), which was similar to that of the \blue-dominant subgroup (3) (Fig. \ref{fig:expert-interview-1}iii). Uncertain about what such similarity means, he used \languagescope to retrieve relevant language cues (Fig. \ref{fig:expert-interview-1}b). The highlighted text from the subgroup 2, \tweet{thoughts and prayers are} clearly expressed \care; on the other hand, the texts \tweet{chicago has lost so many to gun violence} from the subgroup 3 and \tweet{gun law country because they cannot defend themseleves} from the subgroup 10 both concerned the harm but there was a difference in what was responsible for the harm (gun violence vs. gun law). ``{\it [These languages differences] did show the varying aspect of concerns [on this gun issue]},'' but after observing the language cues (Fig. \ref{fig:expert-interview-1}b), he felt he needed to be more cautious in interpreting the ``similar'' language patterns. While he found more evidence for the issue polarization hypothesis, he felt his original set up through comparing the language similarity was insufficient and can be misleading if not inspecting the subtleness of how the languages are used in the issue contexts. He concluded, ``{\it [this tool] offers enough depth and information to allow me to learn from the complexity of messages}.'' 

\begin{figure}[!t]
    \vspace{-1.5em}
    \setlength\tabcolsep{2pt} 
    \begin{tabular}{c}\\
    \includegraphics[width=0.95\linewidth]{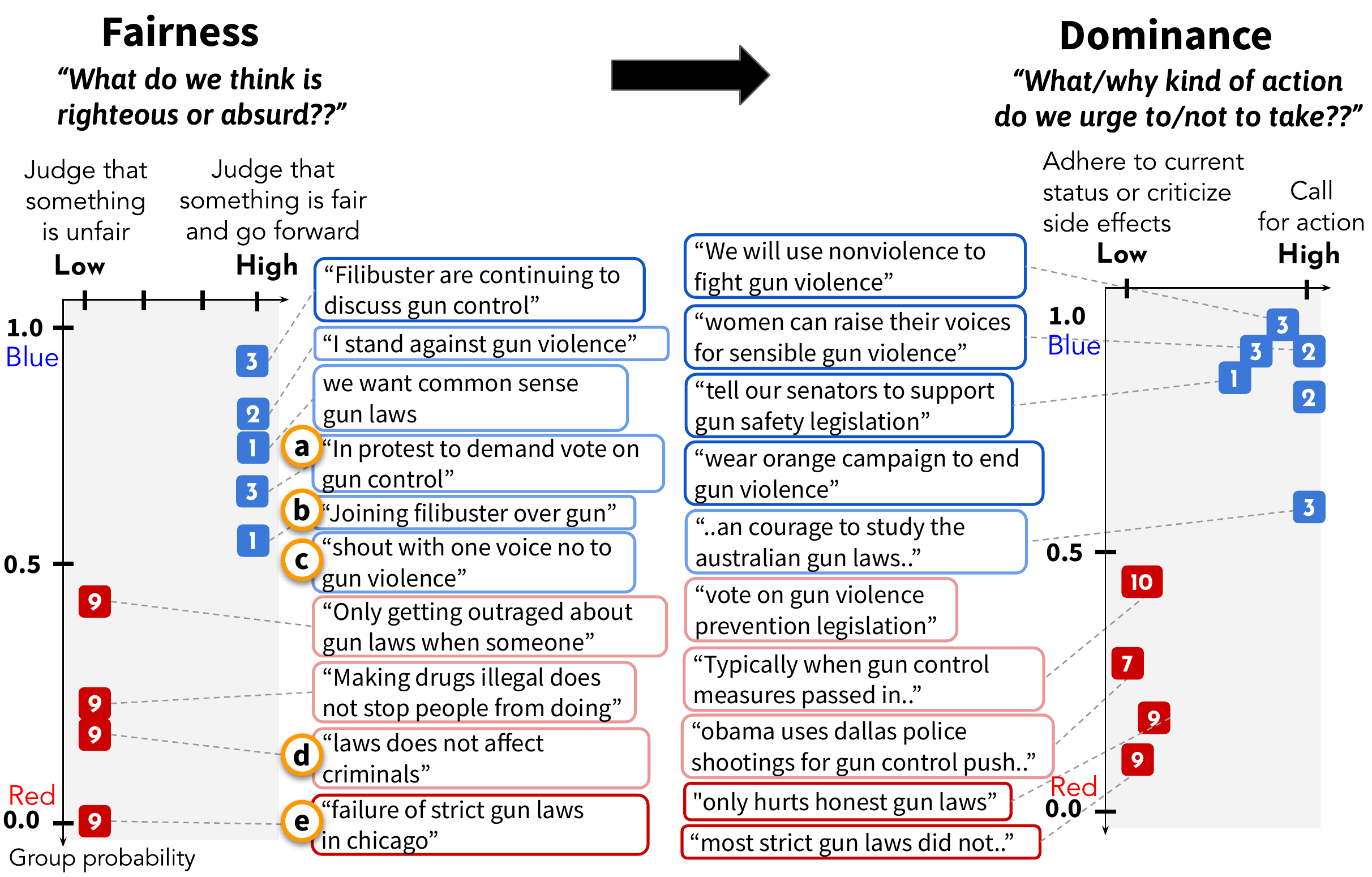}
    \end{tabular}
    \vspace{-1em}
    \caption{\label{fig:expert-interview-2}
    \textbf{Expert interview 2}: Summary of representative language cues for \fair and \domi from \blue-dominated and \red-dominated subgroups. The language cues from the subgroups represent various aspects of online campaigns and debates about guns and gun control policies. }
    \vspace{-1em}
\end{figure}

\subsection{Interview 2: Language insights for online activism and campaigns}

Expert 2 is a social psychologist interested in studying language use and narratives in online movements and campaigns. Having learned about the gun-issue dataset, she was eager to use \name to see how the two political camps differ in psycholinguistic dimensions, particularly in \fair and \domi. She hypothesized the two camps would show different patterns in \domi because she perceived a gradual shift in  public opinions (with recent polls showing increasing support for gun regularization policies), and ``{\it in this backdrop, conservatives may express a lower level of feeling in control}.'' Her hypothesis of the difference in \fair came from her understanding of the central argument on both sides: liberals view the gun regulation as a justified means to fairly guard the public safety ({\it fair}), whereas conservatives view the restriction on gun ownership and rights as putting people in danger ({\it unjust}). She was curious about how her hypothesized differences may reflect in the language used in tweets.

\paragraph{\bf Glance over the group patterns (T1), focus on specific attributes and nuance subgroup patterns (T3)}
Her attention was first drawn to \grouptrend (T1), ``{\it so nice…you can see the overall patterns for the two major group of tweets only at a few glances}.'' She further used the \scopecontroller to select the two focal dimensions (by unchecking others). After looking closely, she confirmed that the differences between the two camps were aligned with her initial hypotheses, and meanwhile, she noticed that the distinction in \fair seems to be greater than that in \domi. Observing this, she was now interested in adding more attributes in \grouptrend to examine whether other dimensions may have better distinguishing power than the two she originally focused on, ``{\it [this makes it] easier to inspect which [additional] dimensions could be more useful in differentiating the groups}.'' She noticed that the \blue tweets tend to cluster more closely around higher values in most dimensions, whereas the \red lines spread wider in all the dimensions, which suggests that some tweets from \red camps might be similar to those from \blue. ``{\it [This shows] a more complex picture [of the \red camp]}.'' Observing this, she concluded that the impressions based on the overall patterns may be too overgeneralized. To examine the complexity, she praised \variscope--subgroup for not displaying a simple, dichotomous picture of the two camps but capturing the varying patterns across subgroups, ranging from the most \blue-dominant group, to in-between purple ones, and to the most \red-dominant group.

\paragraph{\bf Establish test validity with language cues, generate a new hypothesis (T5) and help mitigate overgeneralization on group characterization} 
To examine the patterns beyond the dichotomy and to test her hypotheses, she decided to pick subgroups with distinct colors and compared their languages by the \languagescope. ``{\it [It is] so convenient [that it allows for] a quick check on the language sequences from the subgroups.}'' She mentioned that it was usually a complicated and even a tedious process to check the test validity from the natural language signals, and ``{\it a system like yours really facilitates people to navigate more qualitative, complicated messages beyond numbers.}'' From the \languagescope, she found several tweets supported her original hypotheses. For example (as shown in Fig. \ref{fig:expert-interview-2}), texts from \blue-dominant groups (Fig. \ref{fig:expert-interview-2}a-c) mentioning \tweet{to demand vote,} \tweet{join filibuster,} and \tweet{shout with on voice,} expressing a strong \fair tone about righteousness in advocating legal means to make a change, and texts from \red tweets (Fig. \ref{fig:expert-interview-2}d-e) like \tweet{failure of strict gun laws} and \tweet{laws not affect criminals} expressing a low level of \fair (unfair or unjust) tone. After looking at the language cues more closely, she pointed out that the languages within the \red-dominant subgroup were less coherent and direct, which matched the previous observations that the distribution of the scores varied more widely among \red tweets. For example, one \red tweet was actually in favor of gun regulation, ``\textit{vote on gun violence prevention legislation.}'' More, after reviewing more closely to the \red tweets at the lower end, she concluded, ``{\it this gives [new] insights too! ... makes me think of a new hypothesis that for \textit{Dominance}, at the lower end, the languages used in \red tweets may be less coherent. They shared less common narratives.}'' Our design -- which offers non-dichotomous exploration, together with the chance to inspect the language patterns and their variations -- enables her to engage in the kind of sense-making regarding within-group variations against over-generalized conclusions.

\begin{figure}[!t]
    \vspace{-2em}
    \setlength\tabcolsep{2pt} 
    \begin{tabular}{c}\\
    \includegraphics[width=0.95\linewidth]{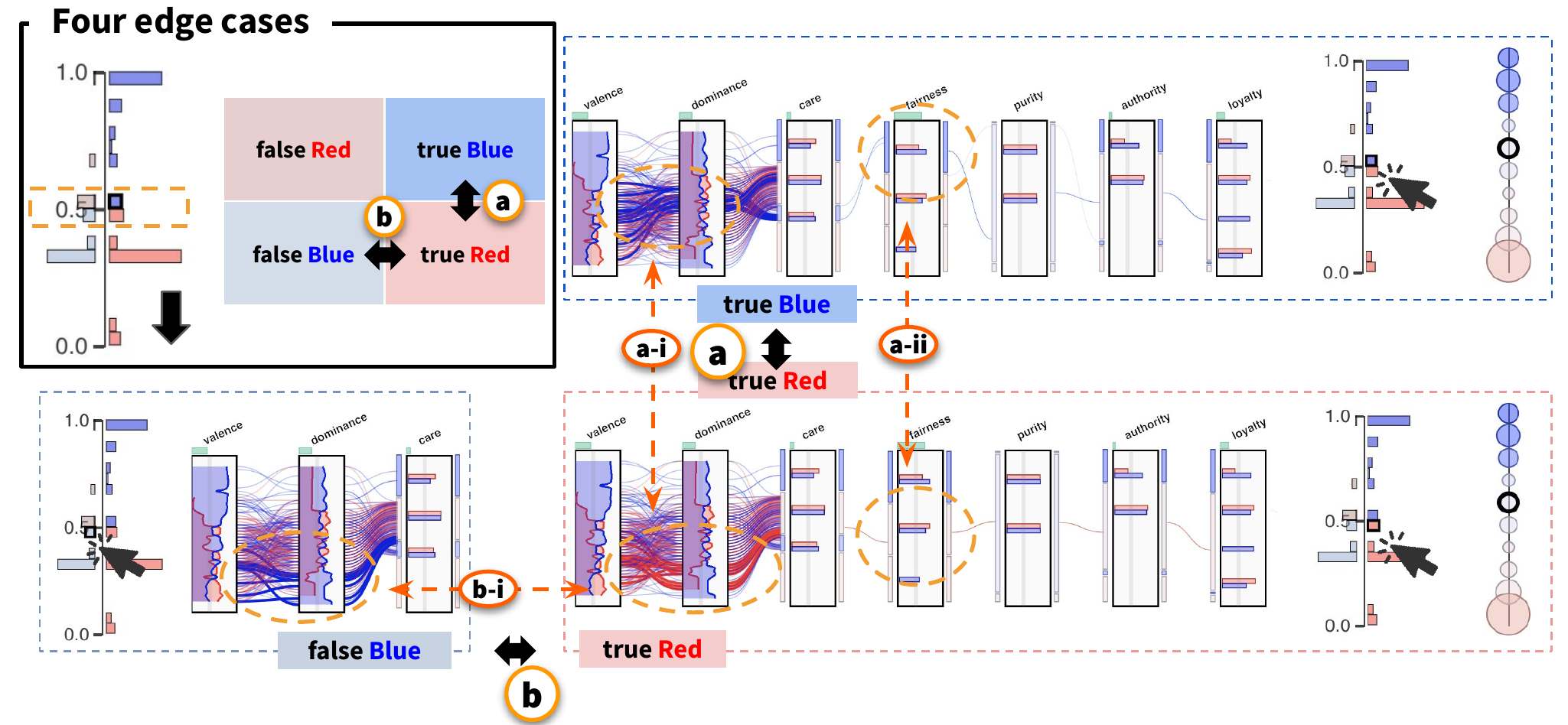}
    \end{tabular}  
    \vspace{-1em}
    \caption{\label{fig:expert-interview-3}
    \textbf{Expert interview 3}: Four edge cases from the subgroup 5 identified through the dual-sided histogram chart in \evalscope. The user selected a set of instances to conduct contrastive analysis: (a) comparing true cases (true \blue vs. true \red), and (b) comparing positive cases (true \red vs. false \blue).
    }
    \vspace{-1.6em}
\end{figure}

\subsection{Interview 3: Interpretable machine learning for discovering common ground and edge cases}

Expert 3 is a natural language processing (NLP) researcher who wishes to 
better understanding the relationship between the interpretable ML's predictions and the groups' psycholinguistic differences.  
While walking through the system, she was particularly interested in examining the subgroup 5, which is a borderline subgroup, having roughly similar portions of members from both camps. She hoped this subgroup might reveal ``{\it what is the psycholinguistic common ground between the two sides?}'' She found \evalscope useful as ``{\it [it] provides an overview of the predictive quality and lets [her] inspect the false and true predictions more closely.}'' She noticed that the subgroup 5 had more edge cases (more false and true predictions with the posterior probabilities close to 0.5), which she thought would bring the interesting finding in understanding the classifier.

\paragraph{\bf Identify edge cases (T2) and examine the inference variability (T3)}
From \grouptrend, she found this particular subgroup had many \blue and \red lines in the middle-ranged values across all dimensions (Fig. \ref{fig:expert-interview-3}). She used \evalscope to find four sets of edge cases and clicked to select each set to see the attribute values across different sociolinguistic dimensions. She found that, when comparing the two true cases (true \blue and true \red on the right), the system gave correct prediction because the differences between the two edge cases, while small, were consistent with the major differences between the two camps -- in terms of \domi and \fair (Fig. \ref{fig:expert-interview-3}a-i and a-ii). When comparing with the two positive cases (false \blue and true \red at the bottom), she found the two sets of instances had similar values of \vale and \domi, but the overall \grouptrend showed that the highlight \blue lines had bigger variance in the two dimensions (Fig. \ref{fig:expert-interview-3}b-i). The lack of coherence in these dimensions from the \blue camp ``{\it would make it trickier for the classifier to do correct prediction.}'' She elaborated that the common ground would be likely to appear from the true edge cases rather than from the incorrect prediction resulted from the noisier basis on either side, and considered the system's ability to tell apart the edge cases very valuable. 

\paragraph{\bf Retrieve similar language patterns (T3) and compare the prediction rationales (T6)}
Noticing some tweets in the two camps had similar sociolinguistic attribute values, she was curious about how the system would explain the differences. She picked two tweets (that have very similar attribute values) and used \rationalescope to check why one tweet was classified as \blue and the other as \red (by selecting the o-mode in the contrastive explanation, as shown in Fig. \ref{fig:teaser}f). She was satisfied when the system returned a rationale indicating \fair as the most discriminative feature for the predictions. She concluded that such level of interpretability that {\it directly links the plain text to sociolinguistic features to group prediction} could be useful to help to determine whether the results were from machine behaviors or human behaviors. 

While using the system, she felt that system helped her gain a better understanding of the interactions between the data and the underlying ML model. She commented that ``{\it the suite of tools allowed [her] to both keep a global view about the data while drilling down to the more interesting subgroups.}'' She was particularly positive about \rationalescope for its contrastive explanation: ``{\it it is useful to be given not just the most discriminative feature but also two contrastive samples; even if I might not personally agree with a particular characterization (say if I don't think this tweet strongly expresses \fair) I at least get a sense for the range that the system is operating under by comparing it against the contrastive tweet.}'' As a suggestion for further development, she anticipate the future system might allow users to construct more directed queries beyond semantic similarity so that a user might dynamically create new data subsets and test new hypotheses.\\

\textbf{Mitigate the overgeneralization on group characteristics (T3-T6).} After exploring functions of \name, she compared it with other relevant tools she had experience with, such as the machine learning tool Weka \cite{witten2016weka}, or the visualization NLPReViz~\cite{trivedi2018nlpreviz} designed for similar purpose. She appreciated the ``analytic engagement'' in the current design -- not simply offering a model concluding what properties may be attributed to a group, but also analytical tools for
users to engage conversations with ``\textit{machine learned classifiers that do over-generalize by its nature and the predictions may not be always correct in general.}'' She explained, ``\textit{I feel that, in the typical group analysis, the burden is on the user’s side. It is usually user’s role to make sure not to jump to conclusion}.'' But, in \name, ``\textit{the overall framework gives the analysts more tools to visualize and study those predictions.}'' She elaborated on the point in details mentioning her experience that, ``{\it Some tools available out there like Weka, for example, offers statistics such as contingency table or uncertainty estimates, which is in the high-level but conveys just one-sided explanations. It does not help break down the reasoning behind it or back up evidences enough to provide the details on identified group characteristics.}'' She particularly highlighted our tool's capability to provide a variety of mechanisms and opportunities to inspect across subgroups at multiple levels of granularity including actual tweet instances and language patterns, so that ``\textit{the analysts could investigate the prediction outcome, the model's underlying rationales, and make informed judgment about whether the hypothesis holds.}'' She added that many tools such as NLPReViz only offered two views, either globally or at individual instances. Finally, in terms of what particular users may benefit from our design, she pointed out,  ``\textit{the contrasting explanation greatly helped non-content experts [who lack prior knowledge of the sociopsychological dimensions] ... it helped them make sense of why the system predicted certain labels}.''

\section{Discussion and Future Work}

\indent \textbf{Discussion.} We discuss the findings and feedback collected during the evaluation with domain experts, and the development process.

\begin{itemize}
    \item \textbf{System utility.} The feedback from the three domain experts are generally positive. The experts mostly agreed that the overall design of \name enabled them to immediately see the trends of each group and subgroup and was useful for testing and refining their analysis hypotheses. The \languagescope was heavily used by Experts 1 and 2 to find qualitative evidence either for supporting existing hypotheses or for generating a more in-depth understanding of how group members may behave differently. The \evalscope and \rationalescope were used more by Expert 3 who concerned the decisions derived from the data and prediction models. Interestingly, all three experts paid significant attention to the ``boundary'' of a group, and the system allows them to check the boundary from different perspectives -- the within-group variability (e.g., \variscope), the language diversity (\languagescope), and the edge cases (e.g., \evalscope).
    \item \textbf{Data generation policy.} Data about people may be generated through certain selection criteria or human coding. For example, our dataset was augmented with human-annotated labels and attributes. However, if the data generation process is not properly communicated, users may misuse or misinterpret the data. We recommend that the data generation policy should be made transparent to the users to the extent possible, and the communication of the data generation process should be incorporated into future design guidelines for accountable group-level analytics.
\end{itemize}

\textbf{Limitation and future work.} Despite that we define our system requirements to be applicable to general scenarios in group analysis, the illustration of our system in this paper is bound to the given dataset and scenario. We discuss the extandability of our system to broader settings of group analysis towards a variety of datasets and applications, and towards multiple groups and social issues. The following paragraphs summarize the limitation and future work with respect to each point.

\begin{itemize}
    \item \textbf{Generalizability.} Our system was demonstrated and evaluated with a twitter data described in Section \ref{sec:appanddata} with seven psycholinguistic attributes, however, we note that our system can incorporate any dataset with a set of attributes in different types (sentiment, topics, behavior, etc.), which are applicable to other domains such as education, business, etc (e.g., inspecting behavioral attributes in team communication). In practice, those attributes can be either manually annotated or automatically derived by automatic methods such as keyword count, latent representation, topic modeling, or log data. 
    While our visualization design is generally applicable for group analysis on text datasets, the language analysis component in the analytic pipeline has limitations. Our system assumes annotations were adequately generated along with the text corpora. Generating proper annotations for different kinds of text corpora -- including texts from various domains such as law, medical, and education -- is beyond the scope of this research. In this work, we use social media texts (tweets written in English) to demonstrate our framework, and thus the language models in our analytic pipeline are trained to have an optimal performance to process social media texts alike. Since the language models are sensitive to the text input, we recommend that the language models should be re-trained and tested based on the text input to ensure the best performance for different kinds of text corpora. Specialized text corpora that require more sophisticated natural language processing modules (e.g., sentence parsing, argument understanding) are also beyond the scope of this work.
    \item \textbf{Scalability.}
    The current design of \name is capable of summarizing group-level patterns, which can be considered as a way of information reduction from large dataset. Nevertheless, the current design is not scalable when exploring a larger number of instances and attributes. In our experiments to test the scalability with number of instances increasing from 3,000 to 20,000, we found that the system experienced degradation in rendering performance with more than 15,000 visual elements. The dataset we use in this paper has 3000 instances. This limited scalability can be potentially improved by using visual aggregation techniques such as edge bundling methods~\cite{holten2006hierarchical}. In case of the large set of attributes, we found that a dataset with more than 20 attributes does not allow enough room for vertical axes to be placed in the visual space. To cope with the issue, our current design supports users to pre-select a smaller set of attributes, which prevents visualization clutter (e.g., from showing too many horizontal axes in the \grouptrend) so the users can inspect the patterns in a more manageable way. 
    Future work can consider incorporating feature selection techniques with filter methods~\cite{guyon2003introduction, chandrashekar2014survey} to help users identify the most interesting set of attributes given appropriate criteria.

    \item \textbf{Perceived reliability.} As described in the design guideline in Section \ref{sec:goalsandtasks}, our tool not only aims to interpret the group difference but supports examining the quality of the model. In the system, \evalscope, which encodes the predictions being rendered as bipolar chart, allows users to observe the distribution of predictions and interactively examine the instances, e.g., whether instances predicted as certain group have particular characteristics. This feature was highlighted by one of the expert interviewee (see Section \ref{sec:expert}) as a novel capability for users to calibrate their confidence of the machine learning model.
    Despite the novel feature, users with different knowledge about machine learning models may see the reliability of the model outcome differently, e.g., users may over-trust a model or overlook the statistical details of the model performance~\cite{nothdurftProbabilisticHumanComputerTrust2014, berkovskyHowRecommendUser2017}. Future work should examine how such ``perceive reliability'' impact the visual analytic system design.
    \item \textbf{Multiple groups.} The current design was optimized for contrasting the trends between two groups. This can be extended to visualize a few more groups -- e.g., by using the multi-color scheme, or using a dichotomous color scheme to generate the one-versus-all comparison. However, the representation of the boundary or edge cases may be not as efficient as that in the current scenario. Future work may explore other visualization and interactive design to help inspect the boundary cases in a multi-group scenario.
    \item \textbf{Multiple social issues.} In our current implementation, we only focus on a single social issue -- the gun-control debates. Future work may look at the group differences across multiple social issues, which creates another level of complexity for exploring the language and attribute distributions within and across groups. One may incorporate approaches such as dynamic queries (interaction), topic modeling (data mining) or hierarchical representation (visualization) to reduce complexity.
\end{itemize}
\section{Conclusion}

In this paper, we proposed \name, a visual analytic system for group differences. Our work is a first attempt at creating a data visualization that aims at promoting a conscientious, interactive experience for users to negotiate with and ponder about analytical results from computational predictive models. The challenge resides in how to retain the complex statistical results to a level that could indicate the group difference patterns derived from computational models succinctly, but not conclusively. Our interface design affords the users opportunities to engage in further analytical thinking beyond what the computational models have offered. Our evaluation by expert interviews suggests \name is a promising design for hypothesis generation and testing for data analysts. 



\section*{Acknowledgement}
We thank the anonymous referees for their useful suggestions. The authors would like to acknowledge the support by the grants from the PICSO Lab, including DARPA UGB, NSF \#1739413, \#2027713, AFOSR awards, and Adobe Research Grant. Any opinions, findings, and conclusions or recommendations expressed in this material do not necessarily reflect the views of the funding sources.
\section{Appendix}

\subsection{Psycholinguistic Attributes}\label{sec:attr-def}

Drawing upon literature~\cite{graham2009liberals, shepherd2018guns,mendez2017neurology}, we identify seven most relevant sociolinguistic attributes that could potentially predict how two ideological groups talk differently on the gun issues.

Two {\it affect} dimensions:
\begin{itemize}
\item \vale: emotions can range from positive (e.g., pleasant, happy, hopeful) to negative (e.g., unhappy, annoyed, despairing)
\item \domi: emotions can range from the most dominant (e.g., feeling-in-control, influential, autonomous) to the least dominant (e.g, weak, submissive, and guided)
\end{itemize}

Five {\it moral} foundations:
\begin{itemize}
\item \care: the virtue of caring, nurturing, and protecting the vulnerable
\item \fair: the virtue of reciprocal altruism, including justice, rights, and welfare
\item \auth: the virtue of respect for authority
\item \loya: the virtue of being loyal to your identified groups
\item \puri: the virtue of seeing the human bodies as holly temples that should not be contaminated
\end{itemize}

\subsection{Human Annotated Attribute Values}\label{sec:annotation}

The human annotation included two phases: (1) creating reliable coding rules, and (2) coding. In the first phase, a major task is to the create the inclusion criteria for human annotators to identify the language signals that correspond to the theorized attributes in tweets. To do so, we sampled a subset of tweets (10-40\%) for each attribute from the total of 3100 relevant tweets. Through an iterative process, one of our authors who is in the field of social psychology began with open-coding to evaluate how the theoretical constructs and categories can be manifested in tweets' language use. She identified the discourse features and themes, and then built, tested, and refining the rules with a graduate research assistant. The inclusion criteria were created with 100\% agreement between the two criteria developers. Once the criteria were set up, each of these tweets was then coded by two independent annotators by a group of four research assistants who did not participate in the the criteria development stage but were trained to follow the coding schemes. These research assistants were chosen because they had been trained prior to this project and developed skills to analyze social media discussions that involve complex politics and social contexts. For each tweet, the annotators determined whether the tweet texts involved each of the seven attributes as a set of binary outcomes. The coding in this phase resulted in fair to substantial agreements between the annotators, with inter-rater reliability in terms of the Cohen's kappa ranging from 0.32 to 0.88 across all attributes. Any disagreement was reconciled after discussion and the coding criteria and procedure were formulated through the process. In the second phase, every tweet (from the 3100 relevant set) was annotated. For {\it moral} attribute values (e.g., \fair, \auth), we followed the coding schemes developed from the first phase. The annotation generated categorical values for each of the moral attribute. For {\it affect} (e.g., \vale), we determined to adopt the Best-Worse Scaling used by Mohammad et al.~\cite{mohammad2018obtaining} after testing it in the first phase. This annotation scheme employed comparative annotation method, which can be used to generate continuous rating for an attribute. We adopted this method and implemented the coding through crowdsourcing on Amazon Mturk. In the crowdsourcing annotation, three annotations are required for each of the $2N$ tweet-tuples (where each tuple contains 4 randomly-grouped tweets, and $N=3100$ in our case) in order generate reliable annotation results. Finally, the annotated scores for affect attributes are normalized to range from -1 to 1.

\subsection{Evaluation for the multi-task prediction}\label{sec:holdout}

\begin{table*}[ht]
\footnotesize
\begin{center}
\caption{\textbf{Results of Multi-task Prediction.} We report results of the Attribute Prediction tasks and Group Label prediction tasks. Performance changes of the Multi-task Predictions compared to baselines are reported in parentheses}\label{tb:pred_performance}
\begin{tabular}{c|c|cc}

\toprule
& \bf Attribute Prediction &\multicolumn{2}{c}{\bf Group Label Prediction} \\

& Acc. (CLF) or Pearson \textit{r}(REG)& Accuracy & F1 \\
\midrule

Dominance (REG) & 0.804 (-0.035) & \bf 0.814 (+ 0.009) & \bf 0.809 (+ 0.014)\\
Valence(REG) & 0.768 (- 0.015) & 0.809  & 0.805\\
Harm(CLF) & 0.623 (+ 0.012) & 0.812 & \bf 0.809 (+ 0.014)\\
Fairness (CLF) & 0.68 (+ 0.001) & 0.803 & 0.802\\
Authority (CLF) & 0.742 (+ 0.007) & 0.799 & 0.794\\
Purity (CLF) & 0.975 (+ 0.003) & 0.793 & 0.791\\
Loyalty (CLF) & 0.825 (+ 0.009) & 0.802 & 0.798\\ \midrule
Dominance Baseline & 0.839 & N/A  & N/A \\
Valence Baseline & 0.783 & N/A & N/A \\
Harm Baseline & 0.611 & N/A & N/A \\
Fairness Baseline & 0.679 & N/A & N/A \\
Authority Baseline & 0.735 & N/A & N/A \\
Purity Baseline & 0.972 & N/A  & N/A \\
Loyalty Baseline & 0.816 & N/A & N/A \\ 
\midrule

Group Baseline & N/A & 0.805 & 0.795\\

\bottomrule
\end{tabular}
\end{center}

\end{table*}

We evaluate the multi-task prediction models using a hold-out experiment on the 3100 relevant tweets, where 50\% samples are used for training, 15\% samples for validating, and the remaining samples for testing. 
We consider two types of baseline models: (a) group prediction baseline 1: to predict group labels with all the annotated attributes on sample tweets using standard machine learning method; (b) group prediction baseline 2: to predict group labels with tweet texts on sample tweets using single-task neural network architecture; and (c) attribute prediction baseline: to predict a single attribute value with tweet texts on sample tweets using single-task neural network architecture. 
The group prediction task is evaluated using {\it accuracy} as our dataset is balanced. 
The attribute prediction tasks are evaluated in terms of the {\it Pearson correlation coefficient} for continuous attributes, and by {\it accuracy} for categorical attributes. 
Table~\ref{tb:pred_performance} report performances of all models and baselines.
The performance gain (or loss) of multi-task models compared to the baselines are reported in the parentheses. 
We highlight key observations from the results: (1) For group prediction, our best model achieves accuracy 0.814, outperforming the two group prediction baselines by up to 30\%. (2) For attribute prediction, the performance measures of our models range 0.768--0.804 in terms of \textit{Pearson} correlation (for the two continuous attributes) and 0.623--0.975 in terms of accuracy (for the five categorical attributes), which is very close to the attribute baseline (with only 0.3\% differences on average). 
Such results suggest that our multi-task models can significantly improve group prediction without sacrificing the performance for attribute prediction.

\bibliographystyle{ACM-Reference-Format}
\bibliography{references}

\end{document}